\documentclass{sig-alternate-10pt}

\usepackage{times} % use times font

\usepackage[letterpaper, top=0.875in, bottom=0.875in, left=0.75in, right=0.75in, footskip=0.25in]{geometry}

\usepackage[utf8]{inputenc}
\usepackage{listings}
\usepackage{xcolor}
\usepackage[hyphens]{url}
\usepackage{hyperref}
\usepackage{paralist}
\usepackage{subcaption}

\iftrue
\newcommand{\michjc}[1]{\textcolor{orange}{MikeC: #1}}
\newcommand{\songtao}[1]{\textcolor{blue}{Songtao: #1}}
\newcommand{\srm}[1]{\textcolor{red}{Sam: #1}}
\newcommand{\favyen}[1]{\textcolor{green}{Favyen: #1}}
\newcommand{\ma}[1]{\textcolor{red}{MA: #1}}
\newcommand{\tim}[1]{\textcolor{orange}{Tim: #1}}

\newcommand{\comment}[1]{}
\else 
\newcommand{\michjc}[1]{}
\newcommand{\songtao}[1]{}
\newcommand{\srm}[1]{}
\newcommand{\favyen}[1]{}
\newcommand{\ma}[1]{}
\newcommand{\tim}[1]{}

\newcommand{\comment}[1]{}
\fi

\newcommand{\name}[0]{TagMe}

\title{TagMe: GPS-Assisted Automatic Object Annotation in Videos}

\author{Songtao He, Favyen Bastani, Mohammad Alizadeh, Hari Balakrishnan, \\ Michael Cafarella, Tim Kraska, Sam Madden\\
MIT CSAIL\\
\{songtao, favyen, alizadeh, hari, michjc, kraska, madden\}@csail.mit.edu\\
}

\begin{document}

\maketitle

\begin{abstract}
Training high accuracy object detection models requires large and diverse annotated datasets. However, creating these data-sets is  time-consuming and expensive since it relies on human annotators. We design, implement, and evaluate TagMe, a new approach for automatic object annotation in videos that uses GPS data. 
When the GPS trace of an object is available, TagMe matches the object's motion from GPS trace and the pixels' motions in the video to find the pixels belonging to the object in the video and creates the bounding box annotations of the object. 
TagMe works using passive data collection and can continuously generate new object annotations from outdoor video streams without any human annotators. We evaluate TagMe on a dataset of 100 video clips. We show TagMe can produce high-quality object annotations in a fully-automatic and low-cost way. Compared with the traditional human-in-the-loop solution, TagMe can produce the same amount of annotations at a much lower cost, e.g., up to 110x.

\end{abstract}

\section{Introduction}
Accurate object detection is a core component of many  applications such as autonomous driving~\cite{wu2017squeezedet}, search and rescue~\cite{bejiga2016convolutional}, traffic surveillance~\cite{song2019vision}, and infrastructure monitoring~\cite{dick2019deep}. 
%\ma{add citations}.  
Existing techniques for training high-accuracy object detectors require large annotated datasets for training, and while pretrained detectors exist for many object classes, their accuracy can be improved significantly via domain-spe-cific fine-tuning~\cite{zhao2019object} on datasets that are similar to those observed in deployment (e.g., similar camera angles, resolutions, lighting, and weather conditions). %\ma{add citation}  
Creating these datasets is a major challenge in practice, often requiring people to manually annotate objects in a huge number of images, which is labor-intensive and expensive.

%\ma{Here is a suggested flow for intro from this point}
%\ma{paragraph 2: State goal (automated object annotation), and key observation: in many scenarios, it is easy to obtain annotated location information. Give some examples. Our focus is GPS but the general approach can be used with other methods for obtaining location information, e.g., RFid tags in an indoor scenario. End with high-level question -- how can we leverage annotated location trajectories to automatically annotate objects in video.} 

%\ma{paragraph 3: Present TagMe and briefly explain how it works using Fig. 1. Summarize its benefits/properties.}

%\ma{paragraph 4: Talk about the technical challenges and how TagMe addresses them. I like the way that paragraph "The core of TagME..." starts. But as explained below, you need to say more about what makes the matching problem difficult} 

In this paper, we focus on automatic object annotation in videos. Traditional object annotation solutions often assume that the videos have already been collected (e.g., from the web) and all their data sources are limited to videos themselves. However, we find in many scenarios, we can easily collect additional location information for objects in videos. For example, in outdoor settings, GPS traces from objects such as vehicles and bicycles may be available either from their embedded GPS receivers or from the smartphones of the riders. The availability of this additional location information provides an opportunity to improve video object annotation: when we need to annotate an object in a video (e.g., bounding box annotation), we may use the location information of the object as a hint to automate the annotation task. The question is how can we systematically leverage this additional location information to automate the video annotation task? 

Why might this approach help? Suppose we want to obtain a large number of annotated images of a new car model under a variety of conditions and angles. If we could obtain videos, e.g., from outdoor traffic cameras, containing that car, and we also had GPS traces associated with that car, then we could use these two datasets to produce a series of annotated images of that car. For example, these GPS traces could be provided by a vehicle fleet of the new car model operated by the manufacturer, a dealership, a car-sharing company, etc.

%However, we find there is a huge missing opportunity in video object annotation - when we need to annotate a specific object in a video, if we can use additional information from that object, for example, the GPS trajectory of it, we can reduce the object annotation effort or even make the object annotation task fully automatic.

\begin{figure}[h]
    \centering
    \includegraphics[width=\linewidth]{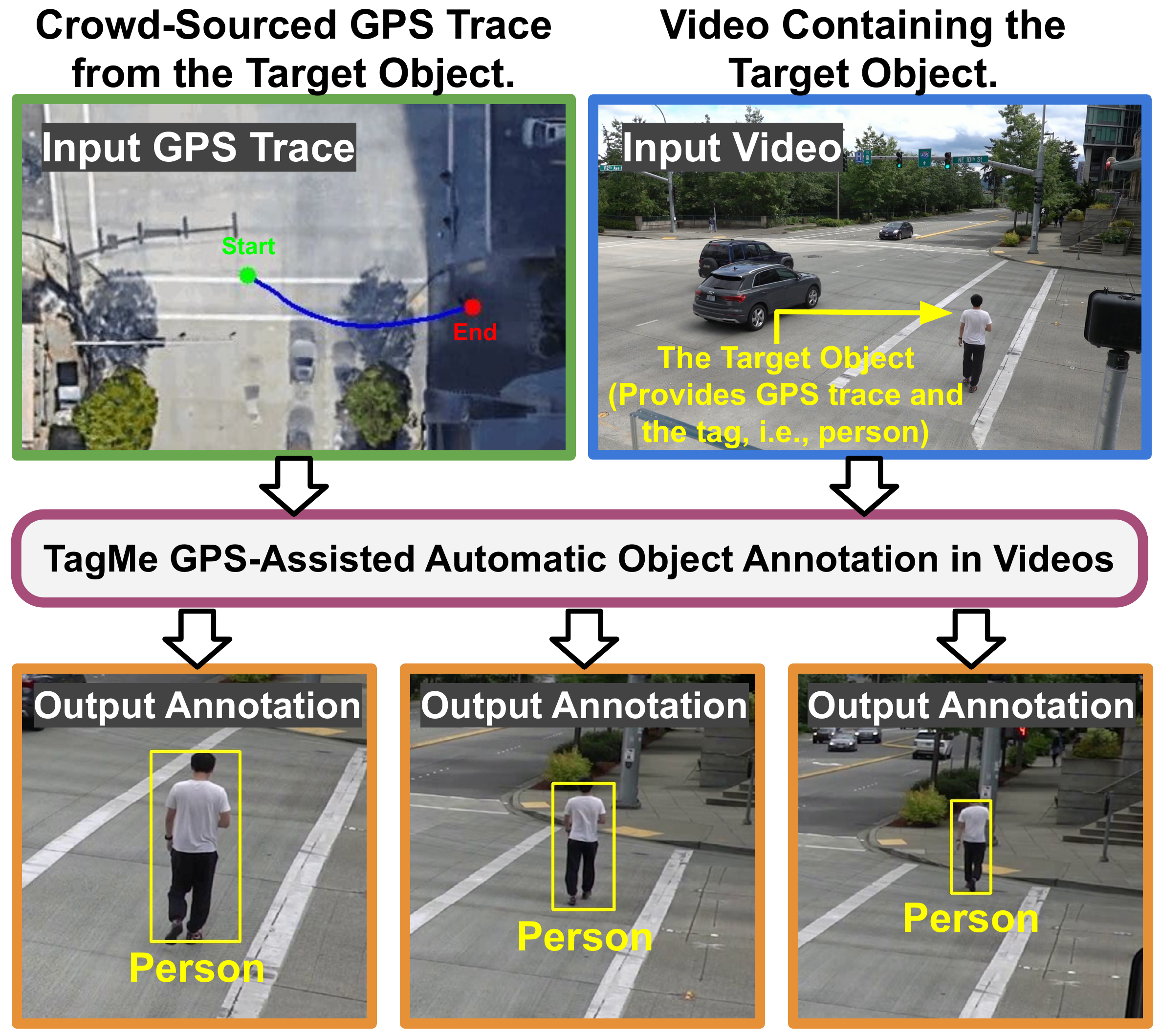}
    %\vspace{-0.05in}
    \caption{Workflow of \name.}
    \label{fig:tagmeintro}
    \vspace{-0.1in}
\end{figure}

{\bf \name}\ is a system that provides this capability. We focus on GPS sensors in outdoor settings for obtaining location traces, but this approach extends to precise indoor localization methods too. We show the workflow of \name\ in Figure~\ref{fig:tagmeintro}. \name\ can automatically annotate objects if their GPS traces are available. When an object (e.g., a person) equipped with a GPS receiver (e.g., a smartphone) is captured by a camera (e.g., a traffic camera), and the corresponding GPS trace is available, \name\ analyzes the motion of pixels in the video and uses the GPS traces to find the most likely pixels belonging to the target object in the video. Once these pixels are found in the video, \name\ produces the bounding box annotation of the target object in each video frame. The whole process is automatic.  

%Once the corresponding pixels of a target object is localized in the video, we can then create the bounding box annotation of the object. The whole process is fully automatic.     

%can use the GPS trace and the corresponding video clip to create bounding box annotations of the object in a fully automatic way.

%In \name, the GPS contributors only need to tag themselves once (i.e., tell the system the category of the object) and other data collection process is transparent to the contributors. Therefore, \name\ can continuously collect annotated dataset at scale. For example, if a car-sharing company wants to create an object detection dataset for different car makes and models, it can use the videos from traffic cameras and the GPS traces from their car fleet to create this dataset. The whole process requires little additional work because the car-sharing company already knows the car models and the GPS traces of their fleet. Because this dataset is collected continuously, it can cover a wide range of diverse scenarios such as different lighting and weather conditions, and different sessions. This diverse dataset can then be used to train high-accuracy detectors for other applications~\cite{MMR}.

%the dataset can cover different times of day, different weather conditions and different seasons. 
%We have already seen many successful use cases like this, for example, Google maps use their users GPS traces to improve the real-time traffic estimation. 

The core problem we solve in \name\ is: given both the video and the location (GPS) trace of the target object, produce the bounding box of the target object in each video frame. A strawman solution to  this problem consists of two steps: (1) detect moving objects in each video frame by anal-ysing the motion of pixels (i.e., optical flows) and (2) map the coordinates of all the moving objects from the frame coordinate to the world coordinate, find the nearest moving object to the GPS position, and use the bounding box of the matched moving object as the output. However, we find that this strawman yields poor precision for three reasons: (1) the moving object detection algorithm is not perfect, e.g., when there are two objects moving close to each other, the moving object detection algorithm may consider them as one object, (2) the GPS trace of the target object is often noisy (when there are multiple objects in the video, the GPS position may be closer to the wrong object), (3) there is no mechanism in this solution to automatically assess the quality of the output annotations, requiring manual verification.

To overcome these challenges, we propose a robust object annotation pipeline with four stages: candidate object proposal, GPS-object matching, bounding box refinement, and bounding box ranking. The key intuition is that the bounding box sequence of an object in the video has to follow some continuity properties; e.g., the sizes and locations of the bounding boxes in two consecutive video frames have to be similar. Our pipeline also uses the continuity properties to rank the annotations, so that \name\ can filter out bad-quality annotations.  

We have implemented \name\ and evaluate it using a dataset of 100 video clips from stationary cameras and UAV drones. Our experiments show that \name{}'s annotation pipeline can improve the precision of bounding box annotations from 28.2\% (the strawman solution) to 88.3\% and 94.4\% if we keep the top-50\% and the top-10\% of the output annotations respectively (through annotation quality prediction and ranking). 
Compared with a human-in-the-loop annotation solution~\cite{googledatalabel}, we find that \name\ can produce the same number of annotations at a much lower cost (all of which is from computing): 110x, 59x, and 18x lower if we keep all, top-50\%, and top-10\% of the output annotations respectively. We also present two case studies to evaluate the end-to-end quality of the generated bounding boxes. We find that when we use the top-50\% of the generated bounding boxes as training data, they have similar end-to-end quality to the manual-annotated bounding boxes in fine-tuning tasks.  

\name\ is an example of using mobile sensing and computing to help with machine learning tasks, in this case automating image annotations using GPS traces as hints, achieving high quality at much lower cost than traditional human-centric methods.

\section{Motivation}
\label{sec:motivation}
In this section, we discuss the motivation behind \name\ system. We focus on three issues, 
\begin{compactitem}
    \item Need for additional object annotations.
    \item Usability of \name's auto-generated annotations.
    \item Use cases of \name\ system in real world.
\end{compactitem}
\subsection{Need for Additional Object Annotations}
Additional annotated datasets are useful for at least two reasons:
%In this work, we aim to find a solution that can provide object annotations in a fully automatic way. In this section, we discuss why we need additional object annotations given many public datasets exist. Here, we list three reasons,%, and why we need to automate the object annotation task.

% \textbf{Why we need more object annotations?} Although there have already been many public dataset available, we still need to label additional data because of three reasons,

(1) \textit{Existing datasets cannot cover all the object categories.} When we need to use an object detector in an application, the first challenge we may face is that there is no dataset covering the objects we want to detect. For example, we may not find a dataset if we want to create an object detector for all the new car models produced this year. In this case, we need additional object annotations to train the model.
%annotate additional images to train a new object detector. 

\begin{figure}[h]
    \centering
    %\vspace{-0.3cm}
    \includegraphics[width=0.8\linewidth]{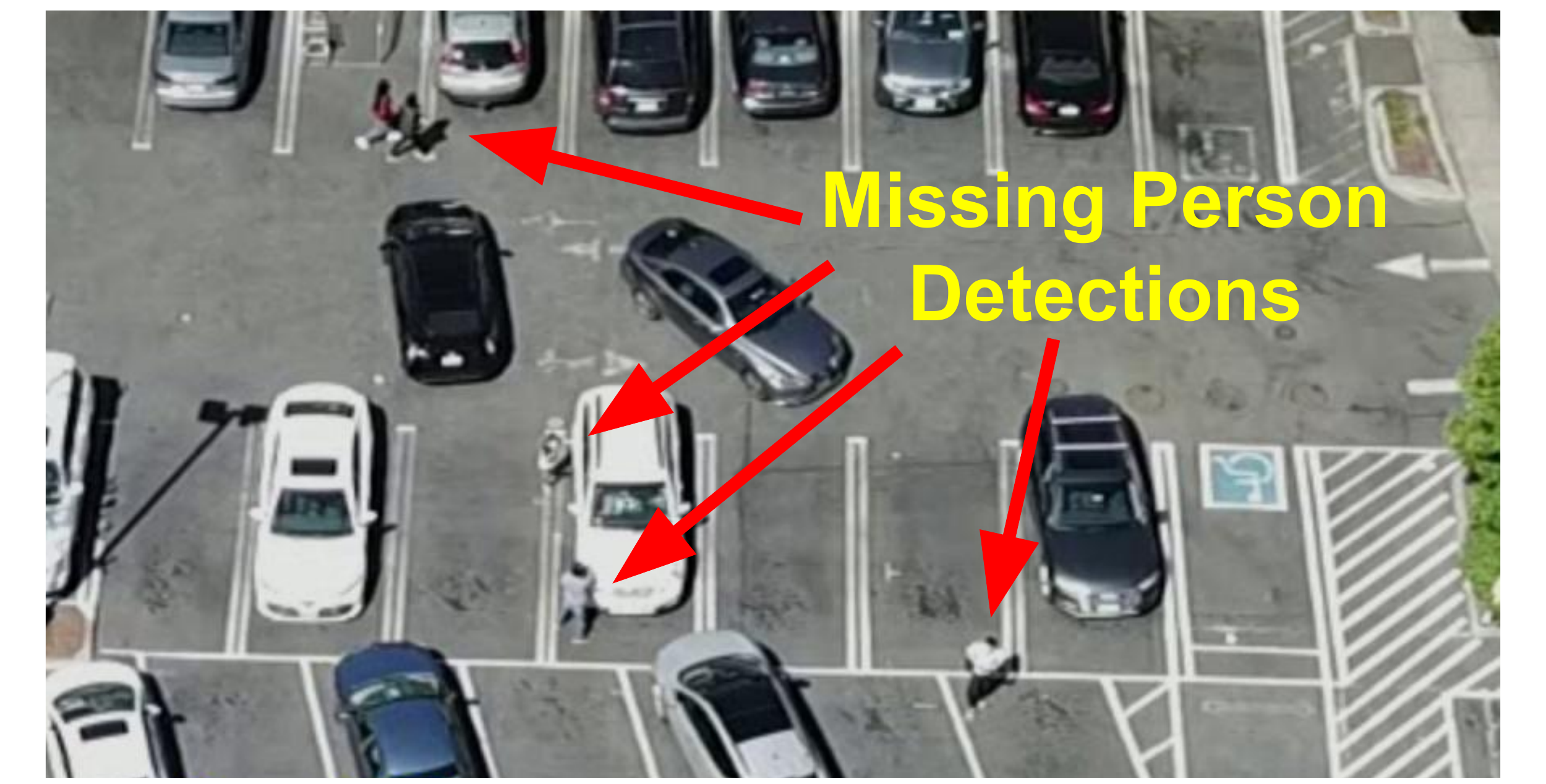}
    %\vspace{-0.2cm}
    \caption{An object detector trained with non-aerial imagery applied to aerial imagery.}
    \label{fig:motivation1}
    %\vspace{-0.3cm}
\end{figure}

\begin{figure}[h]
    \centering
    %\vspace{-0.3cm}
    \includegraphics[width=\linewidth]{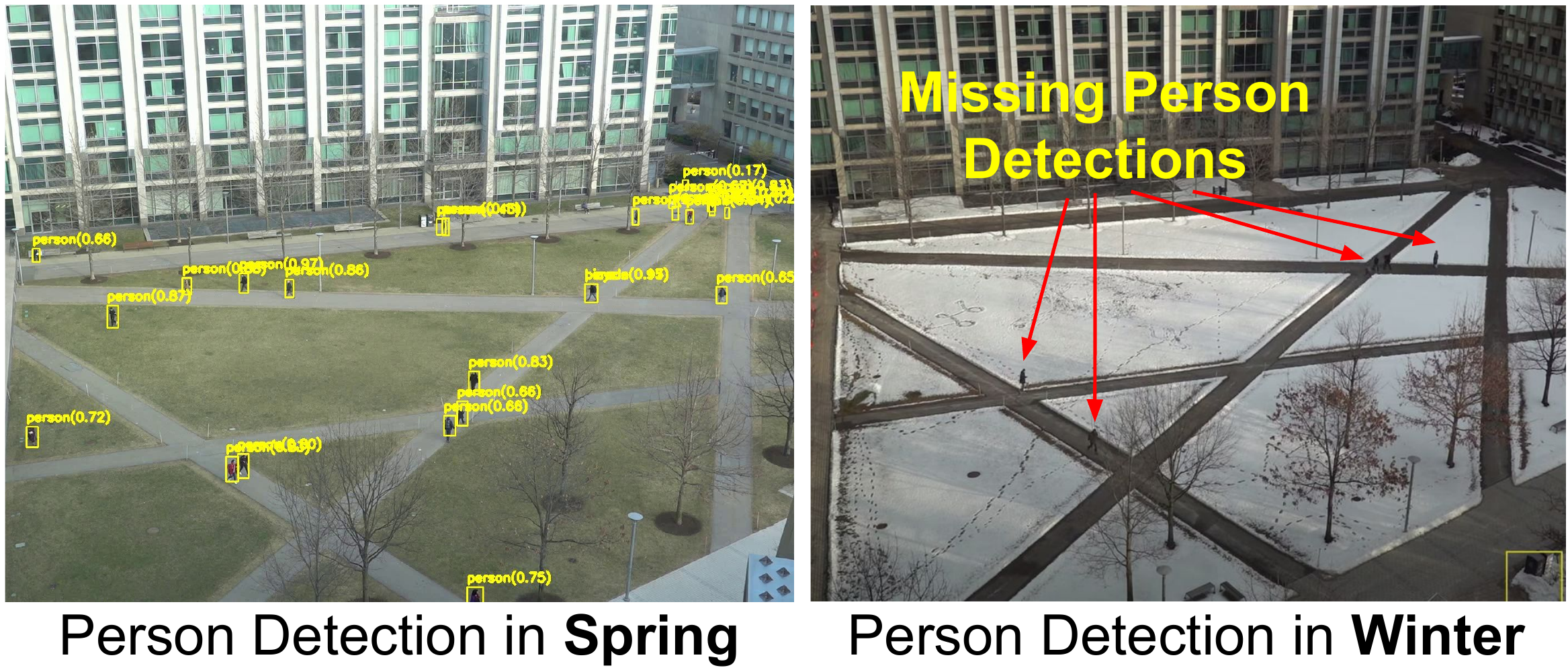}
    %\vspace{-0.2cm}
    \caption{Environment bias of the object detector.}
    \label{fig:motivation2}
    %\vspace{-0.3cm}
\end{figure}

(2) \textit{Scenario biases.} Even if we can find a dataset that covers the objects we want to detect, the dataset may still have scenario biases, e.g., camera-angle bias or environmental bias. For camera-angle bias, we take a YOLO-V3~\cite{redmon2018yolov3} object detector trained on the COCO~\cite{cocolin2014microsoft} dataset as an example. We find that this object detector does not detect persons from top-down aerial images (see Figure~\ref{fig:motivation1}). For environmental bias, the same detector works well for person detection from a camera in spring, but the precision drops in winter (Figure~\ref{fig:motivation2}).

In the above examples, we need to fine-tune an existing object detectors on additional annotated datasets that are similar to the scenarios where the object detectors are deployed.

%(3) \textit{Environment bias.} Another bias we observed in the existing dataset is from the environment. Again, we take a YOLO-V3 object detector trained on the COCO dataset as an example. As shown in Figure~\ref{fig:motivation2}, we find this model works well for person detection from a surveillance camera in spring. However, the accuracy drops significantly in winter. In this case, we may need to annotate additional images collected in winter to fine-tune the object detector.

\subsection{Usability of Auto-Generated Annotations}

\begin{figure}[h]
    \centering
    %\vspace{-0.3cm}
    \includegraphics[width=\linewidth]{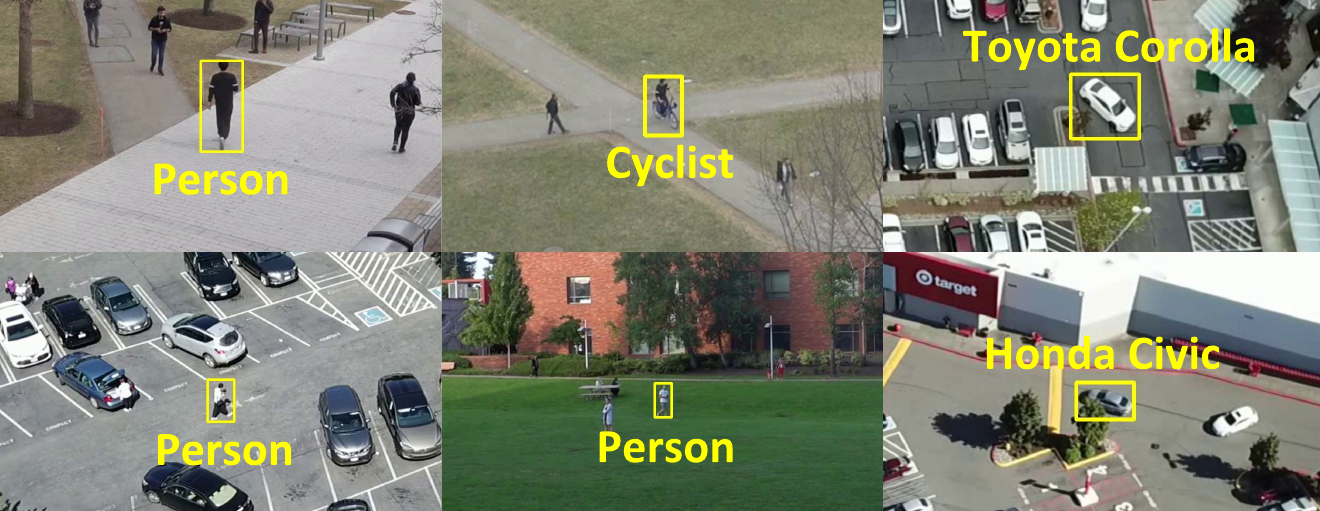}
    %\vspace{-0.3cm}
    \caption{Bounding box annotations produced by \name.}
    \label{fig:samples}
    %\vspace{-0.2cm}
\end{figure}
We show examples of \name's generated bounding boxes in Figure~\ref{fig:samples}. 
%(Appendix~\ref{appendix:qualitative-results}has more examples).
\name\ is capable of producing high-quality annotations. These bounding box annotations can be directly used for image classification models. For object detection models, when there are multiple objects of the same category in a video frame, because \name\ can only annotate objects with GPS traces, we may only have partial annotations for the entire video frame. For example, when the target object is a person, and there is more than one person in a video frame, we will only have the bounding box annotation of the target person. Training or fine-tuning object detectors directly with these partial annotations may yield defective results. Fortunately, this is a known challenge in computer vision and many solutions have been proposed~\cite{missingobjwu2018soft,missingobjxu2019missing,missingobjyang2020object,missingobjzhang2020solving}. In our case study 2, we show an example of how we can use partial annotations to fine-tune a pretrained object detector (Section~\ref{sec:casestudy2}). Although existing solutions may not be sufficient to fully solve the partial annotation problem, we believe more advanced solutions will be developed in the future, perhaps atop annotation systems like \name. 

\subsection{Use Cases for \name\ }

We discuss two use cases: crowd-sourcing scenarios and private scenarios.

\textit{(1) Crowd-sourcing scenarios.} In crowd-sourcing scenarios, contributors provide their object tags (which describe what they are) and GPS traces to \name. \name\ combines the GPS traces and the videos from cameras to produce new object annotations. These new object annotations are then be used to improve the object detection or image classification models in different applications such as autonomous driving and smart traffic light scheduling. Finally, the improved performance of these applications gives the value back to the data contributors. 

We have seen great success in many crowd-sourcing applications deployed in real world, for examples, crowd-sourcing live traffic maps~\cite{jeske2013floating} and crowd-sourcing digital maps~\cite{haklay2008openstreetmap} where both of them use crowd-sourced GPS traces. Therefore, we believe \name\ has the potential to grow into another successful crowd-sourcing application that can benefit the society. % and machine learning research.

\textit{(2) Private scenarios.} As a pure annotation technique, \name\ can also be used in many private scenarios, e.g., annotating equipment at a construction site, annotating livestock in a  farm, etc. In these scenarios, \name\ works in a more controlled environment, e.g., no malicious data contributors and no privacy issues.

%In summary... broader impact of this work
%\textbf{Why we need to automate the object annotation task?} 

%In this section, we describe the annotation pipeline of \name. 
\begin{figure*}[h]
    \centering
    %\vspace{-0.3cm}
    \includegraphics[width=\linewidth]{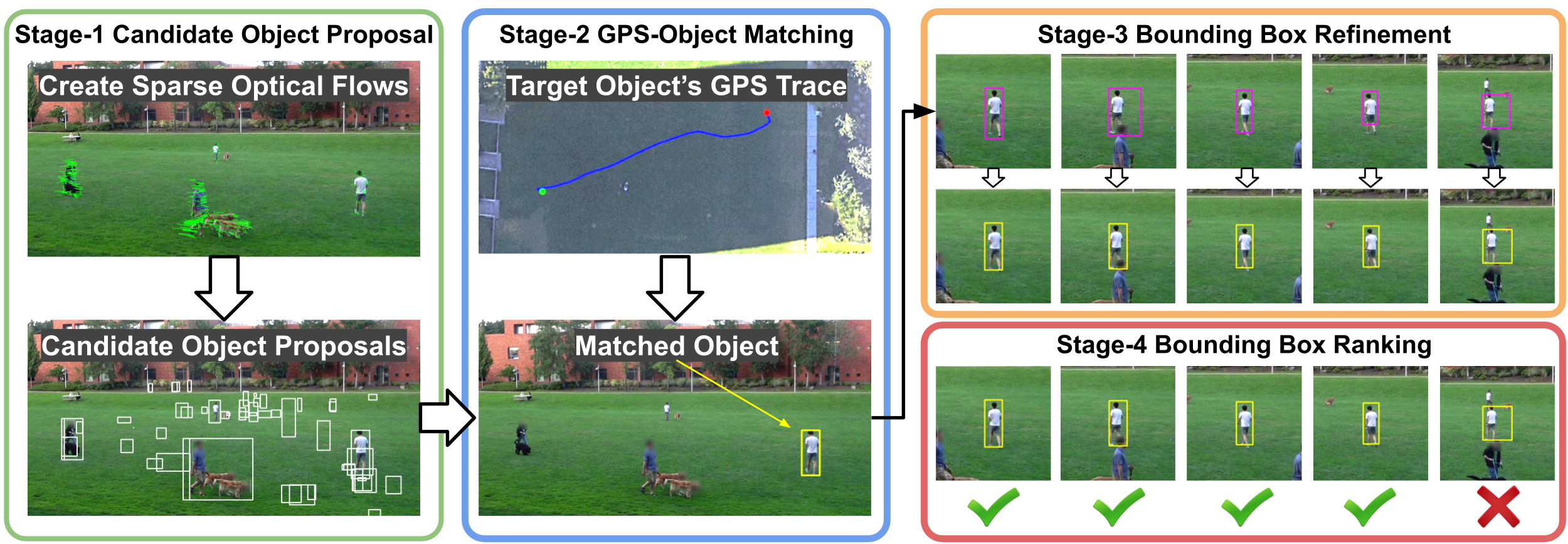}
    %\vspace{-0.1in}
    \caption{Overview of the \name's annotation pipeline.}
    \label{fig:pipeline}
    %\vspace{-0.1in}
    \vspace{-0.3cm}
\end{figure*}
\section{Design}
\label{sec:design}

%\subsection{Overview}
Figure~\ref{fig:pipeline} shows \name's video object annotation pipeline. The goal of this pipeline is to generate the bounding box annotation of a target object in the input video. The pipeline takes both the video and the GPS trace of the target object as input and generates  annotations in four stages:% - (1)candidate object proposal, (2) GPS-object matching, (3) bounding box refinement, and (4) bounding box ranking. 

\textbf{Stage-1: Candidate Object Proposal (Section~\ref{sec:pop}).} The first stage takes the video as input and generates \textit{candidate objects} in each video frame. Each candidate object consists of a group of pixels, allowing us to create the bounding box of the object.
%\ma{perhaps add a sentence about how these bounding boxes are obtained} 
We use these candidate objects as a \textit{superset} of all the real objects in the video, and hopefully, the target object will be in this superset.% We show the details of our candidate object proposal algorithm in Section~\ref{sec:pop}.

\textbf{Stage-2: GPS-object matching (Section~\ref{sec:pom}).} The second stage takes the candidate objects and the GPS trace of the target object as input, and infers the most likely candidate object that matches the target object in each video frame. The matching is challenging because (1) the candidate objects produced in Stage-1 often come with errors such as false objects and noisy bounding boxes, and (2) the noisy GPS locations may be far away from their true locations in the video.
%\ma{A sentence or two about what makes this hard. Be as concrete as possible, For example: Candidate objects from step 1 are noisy (e.g., false objects, missing objects,  misplaced bounding boxes) and  they are inconsistent across video frames (e.g., an candidate object may appear in one frame, disappear in the next, or bounding boxes of objects that are close to each other may overlap or get combined in some frames)} 
%\songtao{Thanks!}
To overcome these challenges, we model this problem as an inference problem in a hidden Markov model (HMM)~\cite{rabiner1986introductionHMM}.  % We show the details of the algorithm in Section~\ref{sec:pom}.   

\textbf{Stage-3: Bounding box refinement (Section~\ref{sec:bbrefine}).} After Stage-2, we get a sequence of candidate objects (one at each frame) and each candidate object comes with a bounding box. 
%So after Stage-2, we get a sequence of bounding boxes that are supposed to correspond to the target object in the video. 
However, because the bounding boxes from the candidate objects may not be perfect, the bounding box sequence may have scattered errors. For example, in Figure~\ref{fig:pipeline}, the generated bounding boxes for the target object are too large in some video frames (e.g., the second column).
%because of the distractions from other nearby objects \ma{distractions from other nearby objects is unclear}. \songtao{I removed this sentence and add more on the source (stage-1) of errors. } 
In the third stage, we fix these scattered errors in the bounding box sequence by exploiting the continuity properties of the bounding boxes using a  sequence-to-sequence neural network model.
%\ma{What is the intuitive reason that the sequence-to-sequence model can do better than whatever approach was used in step 1. Is the idea that having identified the same object in a sequence of frames, we can  use a consistency requirement across frames to refine erroneous bounding boxes?} \songtao{Yes, I rephrased this part to emphasize this intuition.} % We show the details of this model in Section~\ref{sec:bbr}.

\textbf{Stage-4: Bounding box ranking (Section~\ref{sec:bbrank}).} The first three stages generate the bounding box annotation for each video frame. However, some of these bounding boxes may still be inaccurate and should be excluded to produce a high-quality annotated dataset. In the fourth stage, we use another sequence-to-sequence neural network model to predict the quality of each generated bounding box and rank them by the predicted quality scores. This quality ranking allows us to select high-quality (e.g., the top 50\%) bounding boxes.

\subsection{Candidate Object Proposal}
\label{sec:pop}
The goal of the first stage is to find all the potential objects in a video. Object detection in videos is a well-studied computer vision task~\cite{joshi2012survey,yao2019video,ciaparrone2020deep}. Because the goal of this work is to annotate datasets, we adapt an unsupervised approach based on moving object detection to create the candidate object proposals.   

In our approach, we first create the optical flows~\cite{horn1981determining} for the video, which capture the motions of the pixels in each video frame. Here, we use sparse optical flows that estimate the optical flows only for the key points, e.g., the corners. 
%We use OpenCV~\cite{opencv_library} for key point detection and optical flow generation. 

\begin{figure}[h]
    \centering
    %\vspace{-0.3cm}
    \includegraphics[width=0.85\linewidth]{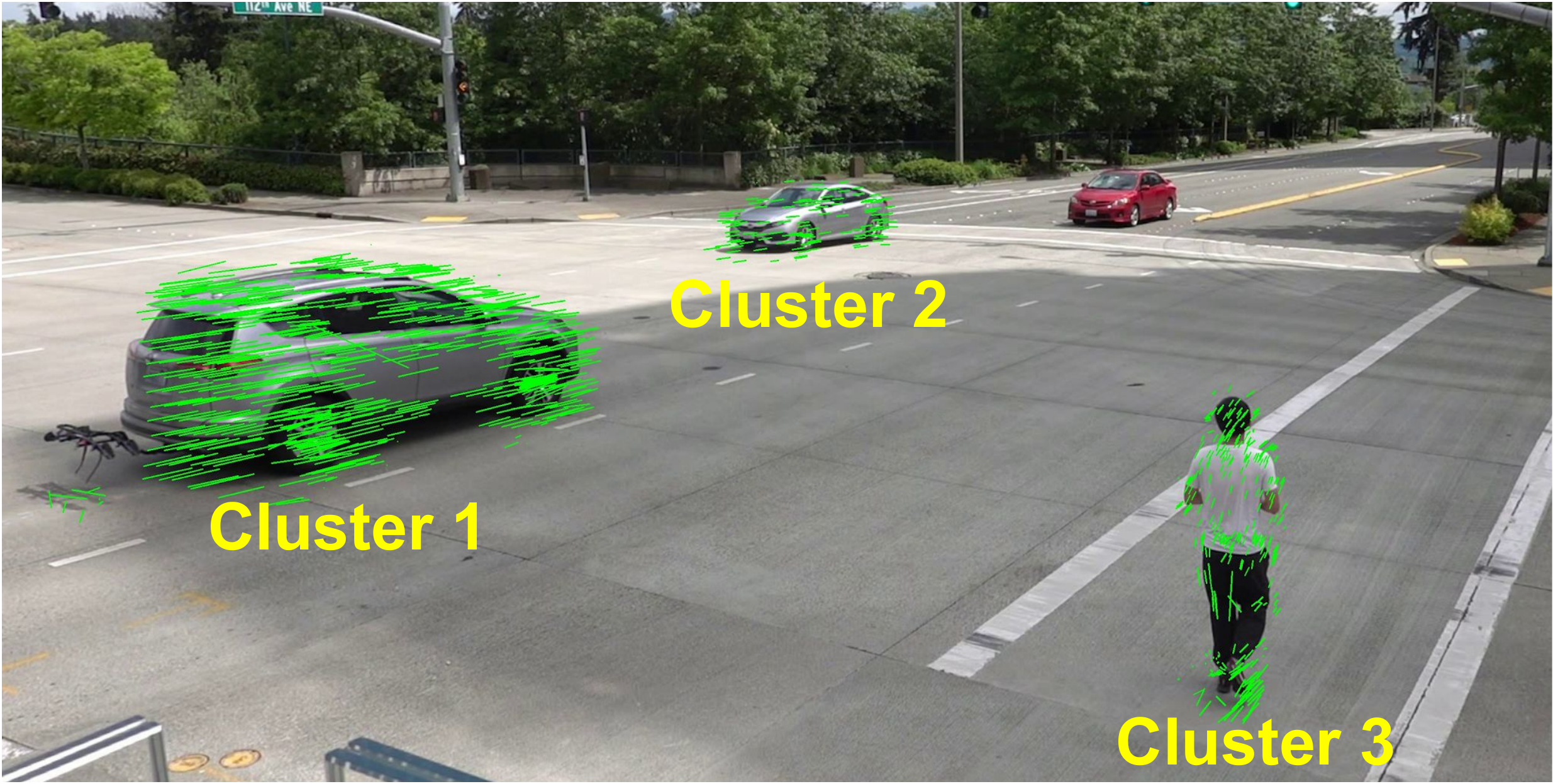}
    %\vspace{-0.1in}
    \caption{We show the moving optical flows of a video frame as green lines. These optical flows form several clusters and each cluster corresponds to a moving object in the video.}
    \label{fig:opticalflows}
    %\vspace{-0.3cm}
\end{figure}

In each video frame (e.g., Figure~\ref{fig:opticalflows}), if we look at all the moving optical flows, we can find that they form several clusters in 2D space, and each cluster corresponds to a moving object in the video. We adapt DBSCAN~\cite{DBSCANester1996density} clustering algorithm to find all the moving objects (clusters) in a video. Once the moving objects are detected, we track each moving object after they have stopped moving to find stationary objects (we also track backward before an object moved). This strategy can cover all the moving objects and most of the stationary objects (they have to move at least once) in each video frame. We call the detected moving objects together with the stationary objects the \textit{candidate objects}. We expect the candidate objects to be a superset of all the objects in a video (e.g., all the white bounding boxes in Figure~\ref{fig:pipeline} Stage-1), and very likely, the target object will be in this superset.    

DBSCAN clustering algorithm has a distance threshold parameter. If the distance between two points is closer than this threshold, the two points are considered connected. We found DBSCAN with a small distance threshold works well for small and far-away objects, while DBSCAN with a large distance threshold works well for large and close-by objects. However, there is no single threshold that can work well for all objects. As the need here is to produce a superset that contains the target object, we choose to trade precision for better coverage. In our algorithm, we use the union of candidate objects produced by DBSCAN algorithm with different distance thresholds as the superset.

%The core of the candidate object proposal algorithm is the clustering algorithm. In \name, we use a clustering algorithm based on DBSCAN~\cite{DBSCANester1996density}. Because the need here is to produce a superset that contains the target object, we design the algorithm to trade precision for better coverage.

%can avoid clustering two nearby objects into one cluster, while at the same time, it can still produce the correct bounding boxes for both small and large objects.  

%Please see Appendix~\ref{appendix:candidate-object-proposal} for further details.   

\subsection{GPS-Object Matching}
\label{sec:pom}
In the second stage, \name\ selects one candidate object in each video frame that is most likely to be the target object. To achieve this goal, \name\ uses the GPS trace collected on the target object as a hint. %Here, we focus on the problem that involves only the GPS sensor. %and briefly show how to extend this to more sensors in Section~\ref{sec:ext}. 

We assume the input videos are calibrated\footnote{We use four points in a video frame to establish a perspective transform between the frame coordinates and the GPS coordinates. We only need to calibrate on one video frame per camera position.} so that we know the GPS coordinate of each pixel in a video frame. For each candidate object, we use the bottom-center pixel of its bounding box to estimate its GPS coordinate. This estimation approach works for most of the objects on the ground.

Once we know the GPS locations of the candidate objects and the GPS location (from GPS sensor) of the target object, we can simply find the nearest candidate object at each frame. However, we find that this basic solution doesn't work because commodity GPS sensors don't have sufficient precision. Meanwhile, the denoising algorithms used in commodity GPS receivers often lead to GPS drifting and motion lags, making it impossible to find the target object from the candidate object pool by just considering the GPS location in one frame.      

\begin{figure}[h]
    \centering
    \vspace{-0.1in}
    \includegraphics[width=\linewidth]{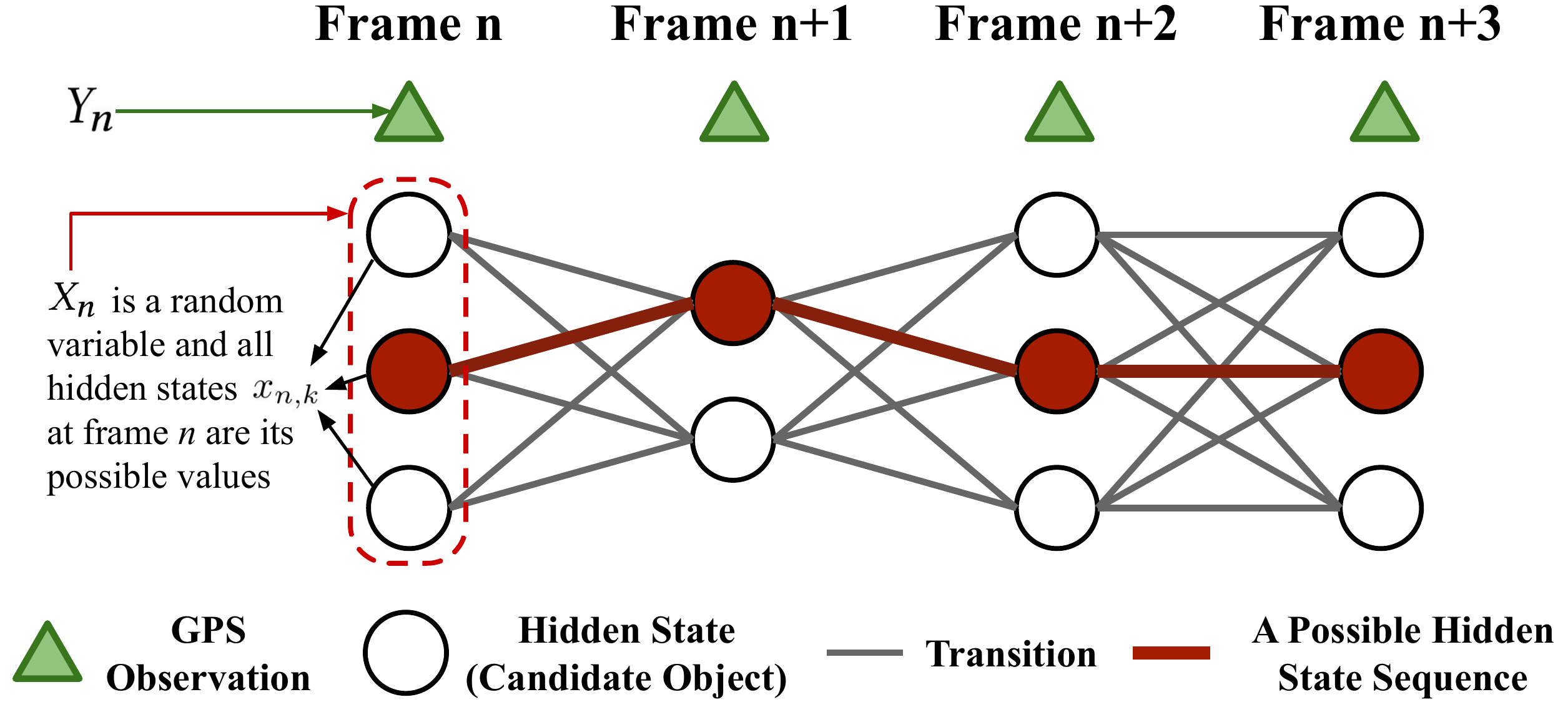}
    
    %\vspace{-0.05in}
    \caption{An illustration of the hidden Markov model.}
    \label{fig:hmm}
    \vspace{-0.0in}
\end{figure}

To solve this problem, we use an HMM. Let $\{Y_n\}$ be the GPS observations from the target object and $\{X_n\}$ be a Markov process representing the sequence of candidate objects that are assigned to the target object in the video. Here, the subscript $n$ corresponds to the $n$-th video frame. $\{X_n\}$ is not directly observable (hidden). As show in Figure \ref{fig:hmm}, in the $n$-th video frame, the candidate objects are represented as hidden states $x_{n,k}$ in the Markov chain. The subscript $k$ represents different hidden states in one frame. For convenience, we omit the subscript $k$ and use $x_n$ instead in the following text. Each hidden state $x_n$ has an \textit{emission probability}, $P(Y_n | X_n = x_n)$, which is the likelihood of observing the GPS location $Y_n$ conditioned on the hidden state $x_n$ being assigned to the target object. In the Markov chain, the hidden state $x_n$ transits to hidden state $x_{n+1}$ following the \textit{transition probability} that depends only on the two involved hidden states $x_n$ and $x_{n+1}$.  

We denote the emission probability of observing the GPS location $Y_n$ at hidden state $x_n$ as $E(Y_n | X_n = x_n)$, and the transition probability from $x_n$ to $x_{n+1}$ as $T(x_n, x_{n+1})$. The GPS-object matching problem may then be stated as follows: Given a video of $N$ frames and $N$ GPS observations $[Y_1, Y_2, ..., Y_N]$, find a sequence of hidden states,  $S = [x_1, x_2, ..., x_N]$ that maximizes the likelihood over the HMM $(\{X_n\},\{ Y_n\})$. 

We can write the likelihood $L(Y | S)$ as
\begin{equation}
E(Y_1|X_1=x_1)\prod_{i=1}^{N-1} T(x_{i},x_{i+1}) E(Y_{i+1}|X_{i+1}=x_{i+1}),
\end{equation}
and the solution to the GPS-object matching problem is 
\begin{equation}
\label{eq1}
    S^* = \operatorname*{argmax}_S L(Y|S).
\end{equation}
The Viterbi algorithm~\cite{forney1973viterbi} solves this problem, but the result $S^*$ may not always match the target object in a video. In fact, the quality of the matching result depends on the emission probability and the transition probability. Next, we show how to construct these probabilities, first as a basic step, and then discuss two improvements.

\subsubsection{Basic HMM} 

We model the emission probability as a 1-D Gaussian function
\begin{equation}
    E(Y_n | X_n = x_n) = e^{-(\frac{d(Y_n, x_n)}{2\sigma_{\text{emission}}})^2},
\end{equation}
where $d(Y_n, x_n)$ is the distance between the GPS observation $Y_n$ and the hidden state $x_n$ (a candidate object). When the distance between the GPS observation and the hidden state is small, the emission probability tends to be $1.0$. This emission probability models the property that the location of the target object should be close to the GPS observation. Here, the scale is controlled by the hyper-parameter $\sigma_{\text{emission}}$.

For the transition probability, there are two possible cases. First, recall that each candidate object (hidden state) is defined by a set of optical flows. If two candidate objects in adjacent frames share a significant amount (i.e., greater than 50\%) of common optical flows, we say these two objects belong to the same object, and we set the transition probability to $1$. Conversely, we model the transition probability as a Gaussian function multiplied by a fixed penalty term. 

We can write the transition probability as
\begin{equation}
T(x_n,x_{n+1}) = \begin{cases}
1.0 & \text{same object}\\
p_{trans}\cdot e^{-(\frac{d(x_n, x_{n+1})}{2\sigma_{\text{trans}}})^2} &\text{otherwise}.
\end{cases}
\end{equation}

This transition probability encourages the transitions between the same object in adjacent frames and penalizes jumps from one object to a different object. 

\subsubsection{Motion Constraints}

For an object in the video, we can link its corresponding candidate object in each video frame to create a sequence of candidate objects. We call this sequence the \textit{candidate object flow (COF)}. Given a COF, we can estimate the object's moving speed and heading angle. Because each candidate object (hidden state) belongs to one COF, we can obtain the speed and heading information for each candidate object (hidden state) in every video frame. Together with the speed and heading estimates from GPS data, we can improve the emission probability by considering more features.

The motion of the GPS data should match the motion of the matched candidate object. We express this intuition in the emission probability as additional \textit{speed and heading constraints}.  
%We propose speed and heading constraints to accurately model the emission probability. The intuition here is that the motions 
%We show the details of these two motion constraints in Appendix~\ref{appendix:motion-constraints}.
%In \name\, to improve the emission probability, we propose two motion constraints, (1)speed constraint, and (2) heading constraint. 

\textbf{Speed constraint.} Intuitively, if the GPS location is moving rapidly, we should not match it to a stationary candidate object. We implement this heuristic into our emission probability. When the moving speed of GPS is faster than a threshold $v_{thr1}$, but the moving speed of the candidate object is slower than a threshold $v_{thr2}$, we set the emission probability to $0$. In this case, $Y_n$ cannot match $x_n$. 

In reality, we find the speed estimation of the GPS may lag the motion of the target object. For example, an object stopped moving at timestamp $10$, but the GPS is still moving toward the true location of the object at timestamp $10$ and stopped moving at timestamp $13$ (3 seconds later). When this happens, the GPS observations from timestamp $10$ to $13$ cannot be matched to the target object because the GPS reading is moving but the target object is stationary. To avoid this issue, we use the minimal GPS speed over a period of time, i.e., 5 seconds, to trigger the speed constraint.   

\textbf{Heading constraint.} Similar to the speed constraint, we can also add a heading constraint. If the GPS observation and the candidate object have opposite moving directions, they are much less likely to be matched. While computing the emission probability of a moving GPS-object pair, we first compute the dot product between the heading vectors from GPS and the candidate object. If the dot product is less than a threshold $\theta_{thr}$, we set the emission probability to $0$. We find this constraint is helpful to ensure correctness when other objects are moving closely to the target object with different headings.

\subsubsection{Shape Preference}

Besides the location information, each candidate object also comes with its shape information, i.e., the width and the height of the bounding box. We improve the basic transition probability by taking this shape information into account. Because the target object's shape should not change significantly between two adjacent frames, the transition between two adjacent candidate objects with similar bounding box shapes should be more likely. We express this \textit{shape preference} intuition in the transition probability. % (see Appendix~\ref{appendix:shape-preference}).

Here, we define the \textit{shape-distance} between two bounding boxes as,
\begin{equation}
    d_{\text{shape}}(w_1,h_1,w_2,h_2)=\text{max}(\frac{w_1}{w_2}, \frac{w_2}{w_1}) + \text{max}(\frac{h_1}{h_2}, \frac{h_2}{h_1}) - 2.0    
\end{equation}
, where $(w_1,h_1)$ and $(w_2,h_2)$ are the widths and heights of the two bounding boxes. We model the shape transition probability as a 1-D Gaussian function, 
\begin{equation}
    T_{\text{shape}}(x_n,x_{n+1}) = e^{-(\frac{d_{\text{shape}}(x_n, x_{n+1})}{2\sigma_{\text{shape}}})^2}
\end{equation}

We multiply this shape term with the original transition probability to get the final transition probability. Here, we use the hyper-parameter $\sigma_{\text{shape}}$ to control the significance of the shape transition term.

\subsubsection{Out-of-Frame Objects}

The target object may not always be within the video frame. It may move into the video frame at the beginning of a video clip and move out of the frame at the end of the clip. This behavior violates the assumption we made in the HMM. To solve this problem, we introduce additional hidden states locating at the boundary of the video frame. These special hidden states can transit to or transit from other normal hidden states (the candidate objects). We find that this simple mechanism works well in practice across our video datasets. In our evaluation, we use this mechanism as a default feature.

\subsubsection{Hyper-parameter Optimization}

In the emission probability and the transition probability, we have defined several hyper-parameters. We don't assign values to these parameters a priori, but search for the best parameters using Hyperopt~\cite{hyperoptbergstra2013making,hyperopt}.

%More specifically, in our evaluation, we use three-fold cross-validation. We split the dataset into three even pieces and search for the best HMM parameters on two pieces and test on the other one. We repeat this process three times to cover the whole dataset. 

\subsection{Bounding Box Refinement}
\label{sec:bbrefine}

\begin{figure}[h]
    \centering
    %\vspace{-0.3cm}
    %\vspace{-0.1in}
    \includegraphics[width=\linewidth]{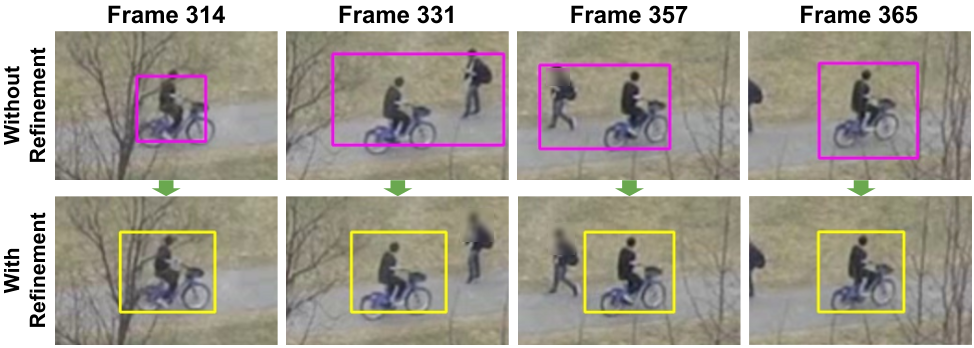}
    %\vspace{-0.15in}
    \caption{Example of bounding box refinement.}
    \label{fig:refine1}
    %\vspace{-0.1in}
    %\vspace{-0.3cm}
\end{figure}

After the GPS-object matching stage, we get a sequence of bounding boxes that are supposed to be the bounding boxes of the target object. However, the bounding boxes may not always fit the target object perfectly. We show an example in Figure~\ref{fig:refine1}. Here, there is another object (a person) moving closely to the target object (a bicyclist). The candidate object proposal algorithm (in Stage-1) failed to produce two separate bounding boxes for the two objects, but instead created an oversized bounding box surrounding the two objects. As a result, the first two stages failed to produce correct bounding boxes in frames 331 to 357.   

To fix the inaccurate bounding boxes, we use a neural network model. The intuition here is that there is generalizable prior knowledge on how a sequence of bounding boxes from one object look. For example, the bounding boxes should be smooth over time and we don't expect to see the size of the bounding box changes rapidly between adjacent frames.

\begin{figure}[h]
    \centering
    %\vspace{-0.3cm}
    %\vspace{-0.1in}
    \includegraphics[width=\linewidth]{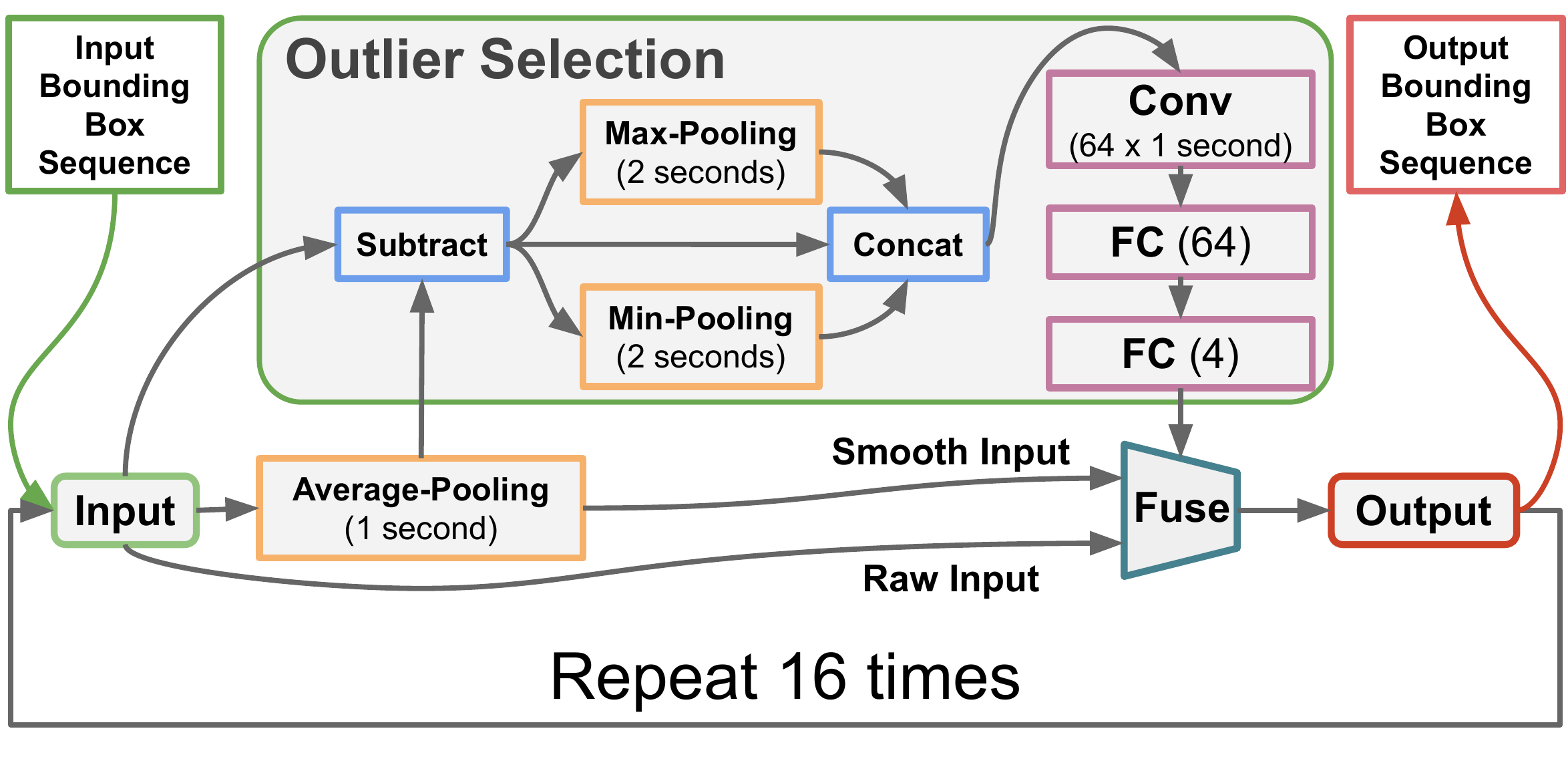}
    \vspace{-0.1in}
    \caption{Bounding box refinement model.}
    \label{fig:stage3model}
    %\vspace{-0.1in}
    %\vspace{-0.3cm}
\end{figure}

We design a sequence-to-sequence neural network model to learn this prior knowledge and use it to refine the bounding box sequence. 
%We show the details of this model in Appendix~\ref{appendix:bounding-box-refinement-model}. As shown in Figure~\ref{fig:refine1}, we find that this model is effective to remove many outliers. 
We show the architecture of our model in Figure ~\ref{fig:stage3model}. The model consists of a learnable outlier filter module. The input to the module is $N$ 4-dimension vectors. Here, $N$ is the length of the bounding box sequence, the 4-dimension vector encodes the center location ($x,y$), the width $w$, and the height $h$ of the bounding box. The outlier filter module refines the bounding box sequences and yields the output sequence in the same format. We pass the output sequence to the outlier filter module again and repeat this process for 16 times to let the outlier filter module gradually refine the bounding box sequence.

Inside the outlier filter module, we first split the input sequence into the raw input sequence $X_{\text{raw}}$ and a smooth input sequence $X_{\text{smooth}}$. We use a convolutional layer followed by two fully connected layers to decide which elements in the input sequence are outliers. Here, we use \textit{second} as the unit to describe the kernel size in a convolutional layer. For example, if the frame rate of the video is 30 FPS, then the size of a 1-second kernel is 30.
We denote the decision from the outlier selection sub-module as $G$, which is a $N$ by $4$ array where each element is a real number between $0$ and $1$. Finally, the output $Y$ is the mixture of $X_{\text{raw}}$ and $X_{\text{smooth}}$ controlled by $G$,
\begin{equation}
    Y = X_{\text{raw}}\cdot(1-G) + X_{\text{smooth}}\cdot G
\end{equation}

We train the entire model with 16 outlier filter modules with $L_1$ loss. Here, we use a relatively simple neural network module and repeat it 16 times rather than using a deeper neural network model to avoid over-fitting.

\subsection{Bounding Box Ranking}
\label{sec:bbrank}

In final state, we predict the quality of each generated bounding box. Although \name\ can continuously generate bounding box annotations from videos and GPS traces, we may not have to use all of them but can select a subset of high-quality bounding boxes. This is because even if we only create annotations for 1\% of the video frames, that will be a huge amount of annotations per day. In order to select high-quality annotations, we need to determine which bounding boxes have better quality compared with others.

To achieve this goal, we use another sequence-to-sequence neural network model
%(see Appendix~\ref{appendix:bounding-box-ranking-model})
to predict the quality score for each bounding box. We can use this score to rank all the bounding boxes, or even rank different video clips, using this quality prediction. 

\begin{figure}[h]
    \centering
    %\vspace{-0.3cm}
    \includegraphics[width=\linewidth]{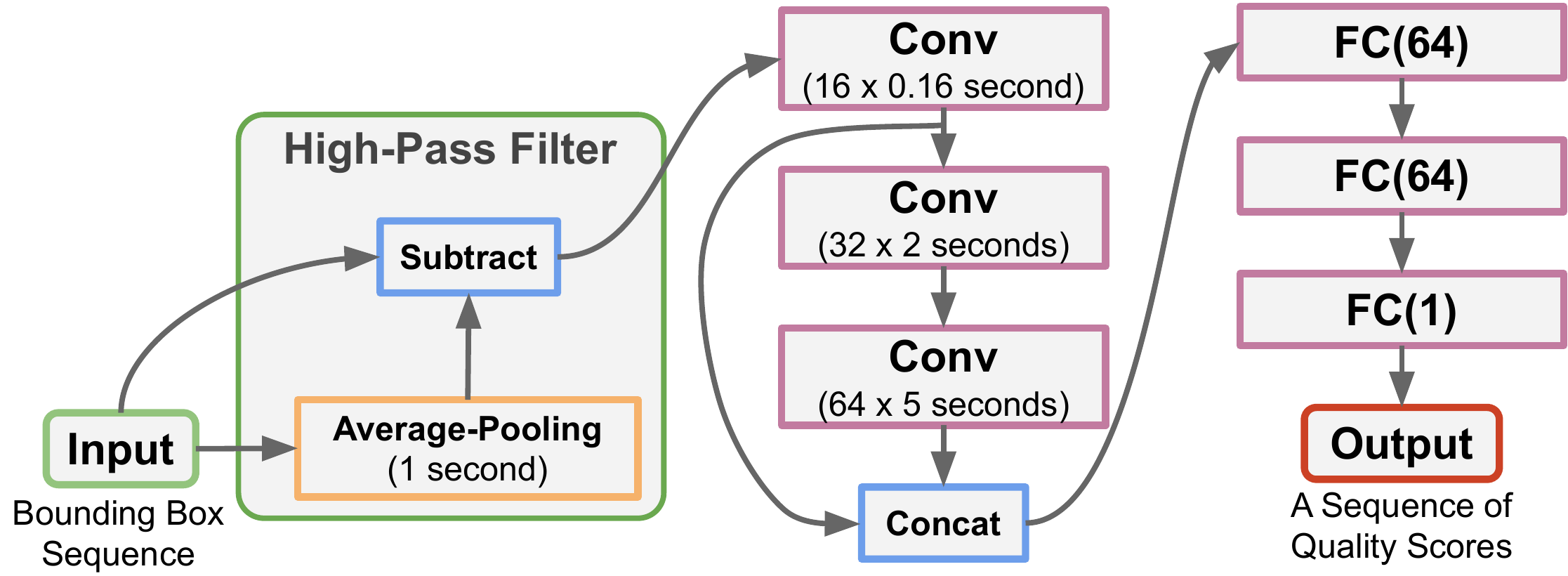}
    \caption{Bounding Box Ranking Model.}
    \label{fig:stage4model}
    %\vspace{-0.2in}
    %\vspace{-0.3cm}
\end{figure}

We show the neural network model in Figure~\ref{fig:stage4model}. The model takes a sequence of bounding boxes as input. It starts with a high-pass filter which is designed to avoid model overfitting. The model consists of three convolutional layers and three fully connected layers. We use a skip connection link to gather information from convolution layers with different kernel sizes. The output score is a real number between $0$ and $1$. We use the IoU of each bounding box as the ground-truth label for the quality score, and train the model with $L_2$ loss. In our evaluation, we find this model can learn good quality scores for ranking propose in both intra-video case (rank bounding box annotations in each video separately) and inter-video case (rank bounding box annotations from all videos together). 

\section{Evaluation}
\label{sec:eval}
\subsection{Dataset}
\begin{figure*}[h]
    \centering
    %\vspace{-0.3cm}
    \includegraphics[width=\linewidth]{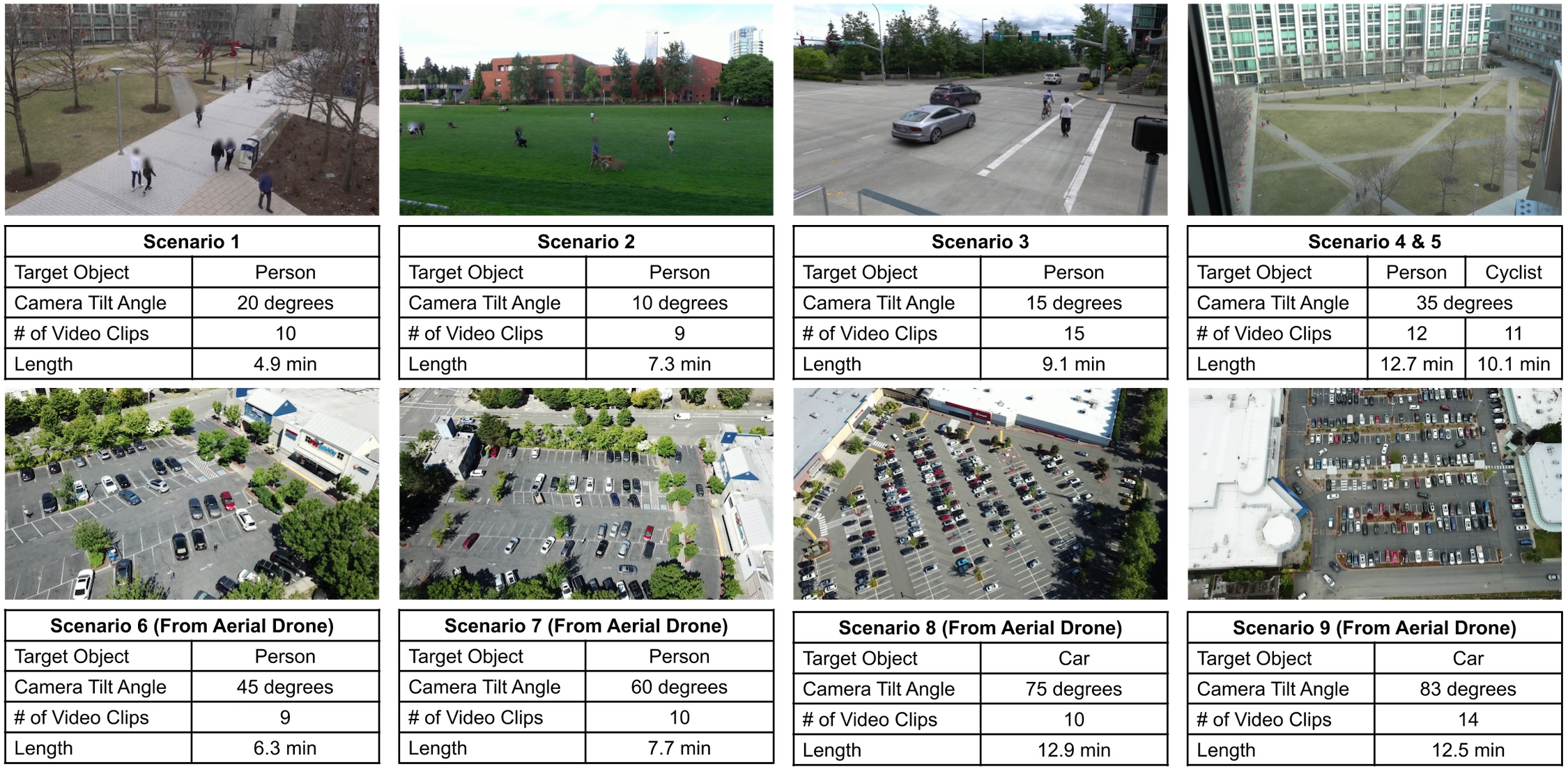}
    
    %\vspace{-0.1cm}
    \caption{Overview of our video dataset. The dataset consists of 100 video clips (30 FPS, 1080p Full HD) from 9 scenarios at 8 different camera positions - 4 of them are from stationary cameras and 4 of them are from aerial drones. }
    \label{fig:dataset}
    \vspace{-0.1in}
    %\vspace{-0.3cm}
\end{figure*}

To evaluate \name, we collected a dataset with 100 video clips from 9 scenarios at 8 different camera positions (see Figure~\ref{fig:dataset}). Each video clip is around 40 seconds long. Along with each video, we collect the GPS traces of the target object at 1 Hz from an Android phone. In this dataset, we use three objects, a person, a cyclist, and a car, as the target objects. The 9 scenarios have different GPS noise levels and varying numbers of other moving objects. The dataset also covers different camera tilt angles from low-tilt angles typical of traffic cameras to high-tilt angles typical of aerial videos. 
%During the data collection, we manually synchronize the clock on the phone and the clock on the video recording devices (i.e., a camera and a drone). 
For each video clip, we manually annotated the bounding boxes of the target object in each frame as ground-truth (the target object appears in 96.5\% of the video frames).

%We use this  dataset to study  the performance and limitations of  \name\ system.

% While we collect the dataset, we manually synchronize the clock on the phone and the clock on the video recording devices (i.e., a camera and a drone) by showing the clock readings from the phone to the video recorders at the beginning of each videos and use the recorded clock readings to synchronize the clocks afterward.

%We show an overview of our dataset in Figure~\ref{fig:dataset}. In the dataset, we use three objects, person, cyclist, and car, as the target objects (all from ourselves, so there is no privacy issue). We collect videos from 8 different camera positions, each of them has different GPS noise levels and different amounts of other moving objects. The dataset also covers different camera tilt angles from low tilt angles which are typical for traffic cameras to high tilt angles which are typical for aerial videos. 

%We intend to use this diverse dataset to give a comprehensive study on the performance and the limitations of our \name\ system.

\subsection{Evaluation Metrics}

We use two metrics to evaluate the quality of the generated bounding boxes. First, \textit{Intersection-over-union} (IoU), which evaluates the similarity between two bounding boxes. Second, the \textit{normalized-distance} (ND), which measures the distance between the centers of the proposed and ground-truth bounding boxes and normalizes this distance by the diagonal size of the ground-truth bounding box. ND evaluates the localization quality of the generated bounding boxes.

\begin{figure}[h]
    \centering
    %\vspace{-0.3cm}
    \includegraphics[width=0.8\linewidth]{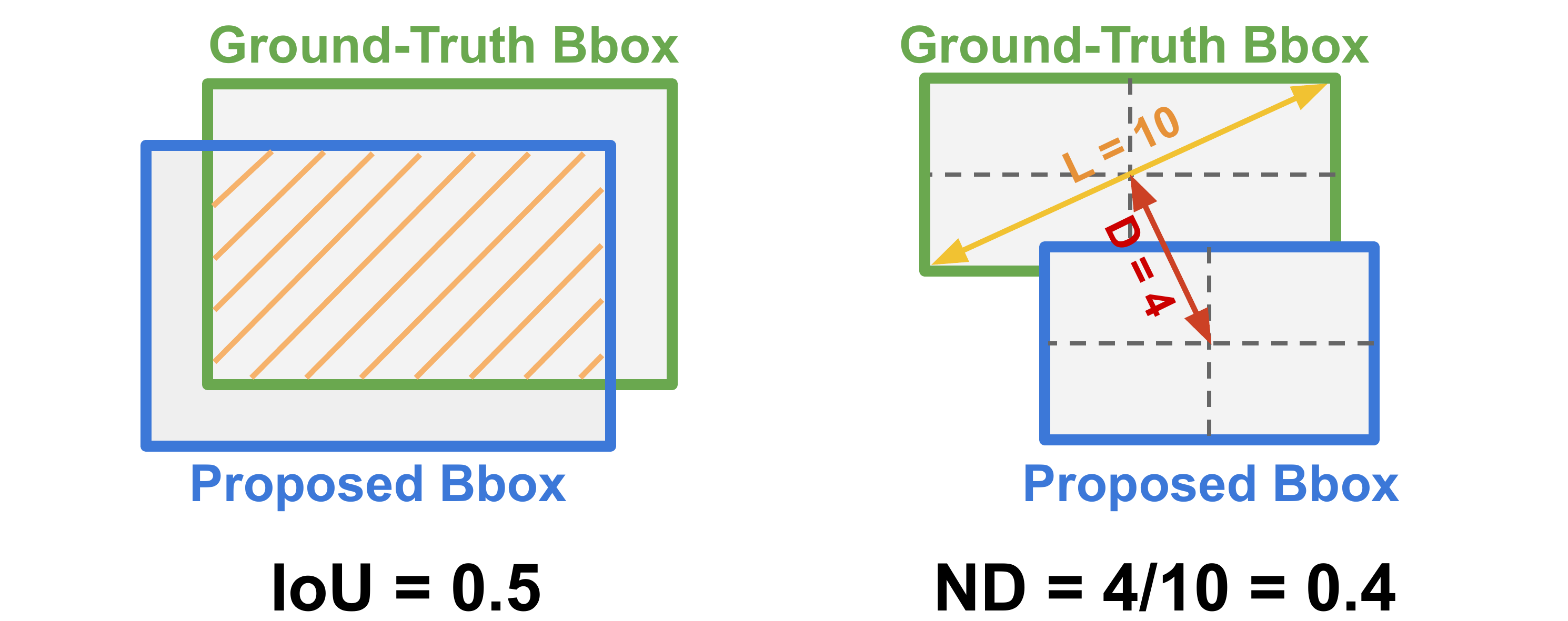}
    \caption{Examples of the IoU metric and the normalized distance (ND) metric.}
    \label{fig:metric}
    \vspace{-0.1in}
    %\vspace{-0.3cm}
\end{figure}

To evaluate a video clip, we use \textit{Precision-at-IoU-0.5} which measures the percentage of bounding boxes whose IoU is greater than $0.5$ in the clip, and \textit{Median-ND}, which measures the median normalized distance for all the bounding boxes in the clip.

\textbf{Three-Fold Cross-Validation.} In all the experiments where hyper-parameter search or model training is needed, we use three-fold cross-validation. We split the dataset into three even pieces and search for the best parameters or train the neural network models on two pieces and test on the other one. We repeat this process three times to cover the whole dataset.

% \subsection{Candidate Object Proposal}
% \todo{remove}
% The goal of the candidate object proposal is to find a superset of objects in each video frame. The coverage of the candidate objects is crucial because the succeeding stages rely on the assumption that the target object is within the candidate objects. We evaluate the quality (coverage) of the generated candidate objects from our method. Please see Appendix~\ref{appendix:candidate-object-proposal} for details.

\subsection{Bounding Box Generation}
The bounding box generation algorithm covers the candidate object proposal stage (Stage-1), the GPS-object matching stage (Stage-2) and the bounding box refinement stage (Stage-3). We first discuss an ablation study of the overall performance of the bounding box generation algorithm, and then discuss the performance and the limitations of the algorithm in different conditions. 

\vspace{0.1in}
\textbf{Ablation Study.} In Table~\ref{table:version}, we show the different versions of the bounding box generation algorithm and their corresponding version names, from the baseline (base) to the proposed algorithm with all features (V4). 
\begin{table}[h]
    \centering
	\begin{tabular}{|l|c|c|c|c|c|} 
		\hline
		Features \textbackslash\  Versions &Base&V1&V2&V3&V4\\ \hline
		Nearest Box & \checkmark & & & & \\ \hline
		Basic HMM & &\checkmark &\checkmark &\checkmark &\checkmark \\
		+ Motion Constraints & & &\checkmark &\checkmark &\checkmark \\
		+ Shape Preference & & & &\checkmark &\checkmark \\ \hline 
		BBox. Refinement & & & & &\checkmark \\ \hline 
	\end{tabular}
	\caption{Versions of the bounding box generation algorithm.}
	\label{table:version}
	\vspace{-0.1in}
	%\vspace{-0.3cm}
\end{table}

%Here, we use type-B IoU, which takes both missing bounding boxes and spurious bounding boxes into account. 

\begin{figure}[h]
    \centering
    %\vspace{-0.3cm}
    \includegraphics[width=\linewidth]{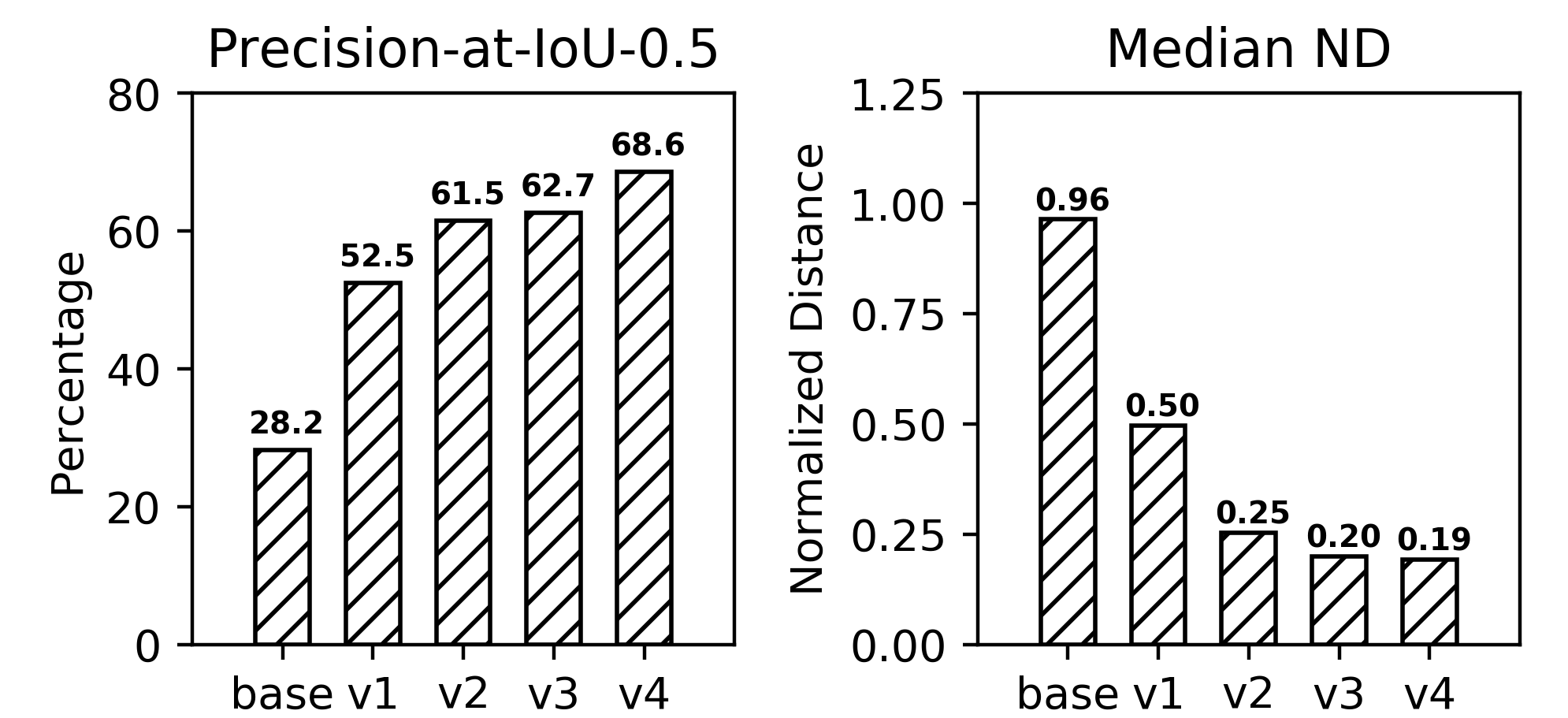}
    
    %\vspace{-0.3cm}
    \caption{Ablation study on different features.}
    \label{fig:stage23bar}
    \vspace{-0.1in}
    %\vspace{-0.3cm}
\end{figure}

In Figure~\ref{fig:stage23bar}, we show the precision-at-IoU-0.5 and the median-ND results for the five different versions. Because of the limited GPS precision, the baseline approach (base), which uses the nearest candidate object as the output in each video frame, does not perform well on both metrics. After improving it with the basic hidden Markov model (V1), we find that the quality of the generated bounding boxes improves significantly as the HMM takes temporal information into account.

We evaluated two improvements (V2 and V3) over the basic HMM. We find that both of them produce additional improvements in terms of the quality of the generated bounding boxes because they introduce more constraints to the HMM and make the probabilistic model more realistic. 

Finally, we use a sequence-to-sequence neural network model to refine the generated bounding box sequence (V4). We find the refined bounding boxes provides much better IoU and a slight improvement in the localization quality (median ND).  

Overall, we made several improvements over the baseline approach and improved the precision-at-IoU-0.5 from 28.2\% to 68.5\% (2.4x) and decreased the median-ND from 0.96 to 0.19 (5x).

\textbf{Performance in Different Environments. }
\label{appendix:performance-in-different-environments}
We show how different factors in the environments may affect the quality of the generated bounding boxes. In this evaluation, we study the following three factors, (1) GPS Error, (2) Count of Moving Objects in the Video, and (3) Camera Tilt Angle.

\textbf{(1) Impact on GPS error.} For each video clip, we measure the GPS error by measuring the distance between the GPS observation and the ground-truth location at each frame. Because we know the perspective transformation from the frame coordinate to the world coordinate, we can estimate the ground truth location in the world coordinate from its bounding box's location in the frame coordinate. 

\begin{figure}[h]
    \centering
    %\vspace{-0.3cm}
    \includegraphics[width=\linewidth]{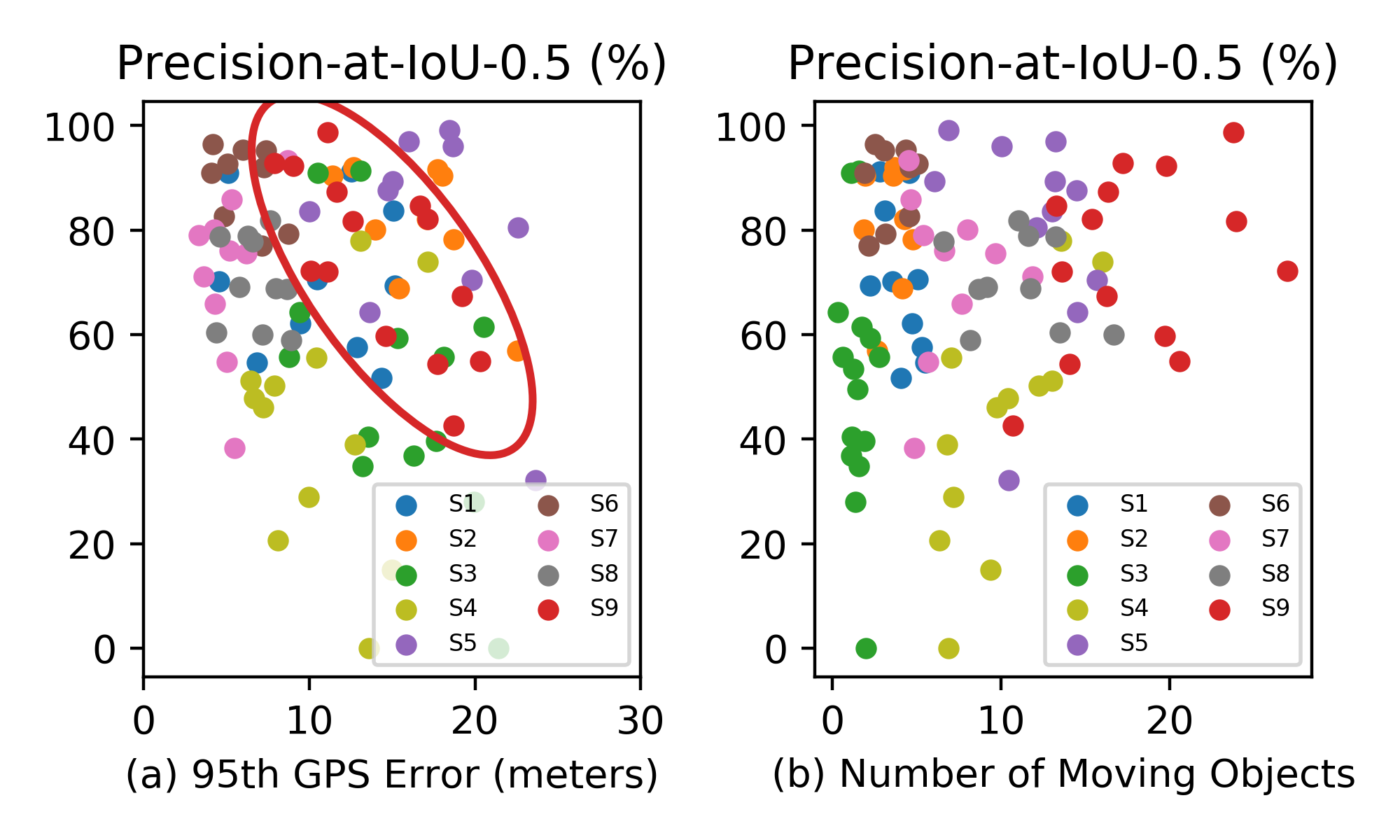}
    %\vspace{-0.3cm}
    \caption{Per-clip Performance vs. the Tail GPS error and the average count of moving objects.}
    \label{fig:stage23colgps}
    \vspace{-0.1in}
    %\vspace{-0.3cm}
\end{figure}

In Figure~\ref{fig:stage23colgps}(a), we show how the GPS error can impact the performance of our bounding box generation algorithm. Here, we report the 95th tail GPS error for each video clip. We color the data points (each data point is a video clip) from different scenarios with different colors.  

\begin{figure}[h]
    \centering
    %\vspace{-0.3cm}
    \includegraphics[width=\linewidth]{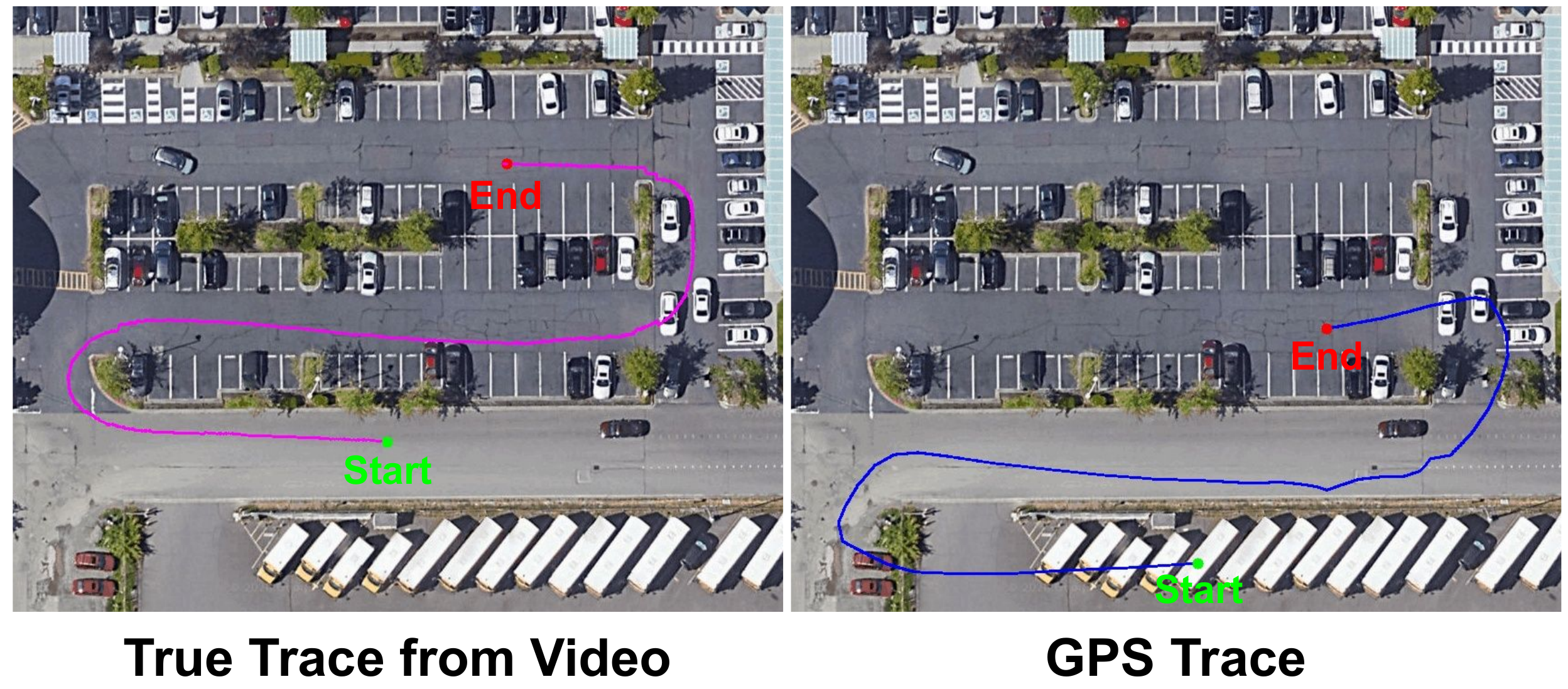}
    %\vspace{-0.1cm}
    \caption{GPS trace may have constant bias.}
    \label{fig:gpsbias}
    \vspace{-0.1in}
    %\vspace{-0.3cm}
\end{figure}

In general, we find there is a weak negative-correlation between the performance and the GPS error. Among all the scenarios, we observe a strong negative-correlation in scenario 9 where there are constant GPS biases in some of the video clips. We show an example in Figure~\ref{fig:gpsbias}. In most of the cases, our matching algorithm can handle this constant GPS bias, however, with the magnitude of the bias increases, the matching algorithm becomes more likely to fail.  

We find in other scenarios, although the GPS errors exist, they are unbiased over time. HMM can handle most of the unbiased errors. Therefore, we observe a week correlation between the performance and the GPS error in most of the cases.

\textbf{(2) Impact on the count of moving objects.} We use our candidate object proposal algorithm to estimate the count of moving objects at each video frame. In Figure ~\ref{fig:stage23colgps}(b), we show how the count of moving objects can impact the performance of our bounding box generation algorithm. 

As our bounding box generation pipeline needs to pick up the target object from all the moving objects in a video, intuitively, the problem becomes harder when there are more moving objects in the video. However, we find our algorithm is robust to the count of moving objects. This is because even though there are many moving objects in the video, they may not all take the same path as the target object does. In this case, the HMM algorithm can still distinguish them.

% \begin{figure}[h]
%     \centering
%     %\vspace{-0.3cm}
%     \includegraphics[width=\linewidth]{Figures/stage23_col.png}
%     \vspace{-0.3cm}
%     \caption{Per-clip Performance vs. the average count of moving objects.}
%     \label{fig:stage23col}
%     %\vspace{-0.3cm}
% \end{figure}

\begin{figure}[h]
    \centering
    %\vspace{-0.3cm}
    \includegraphics[width=\linewidth]{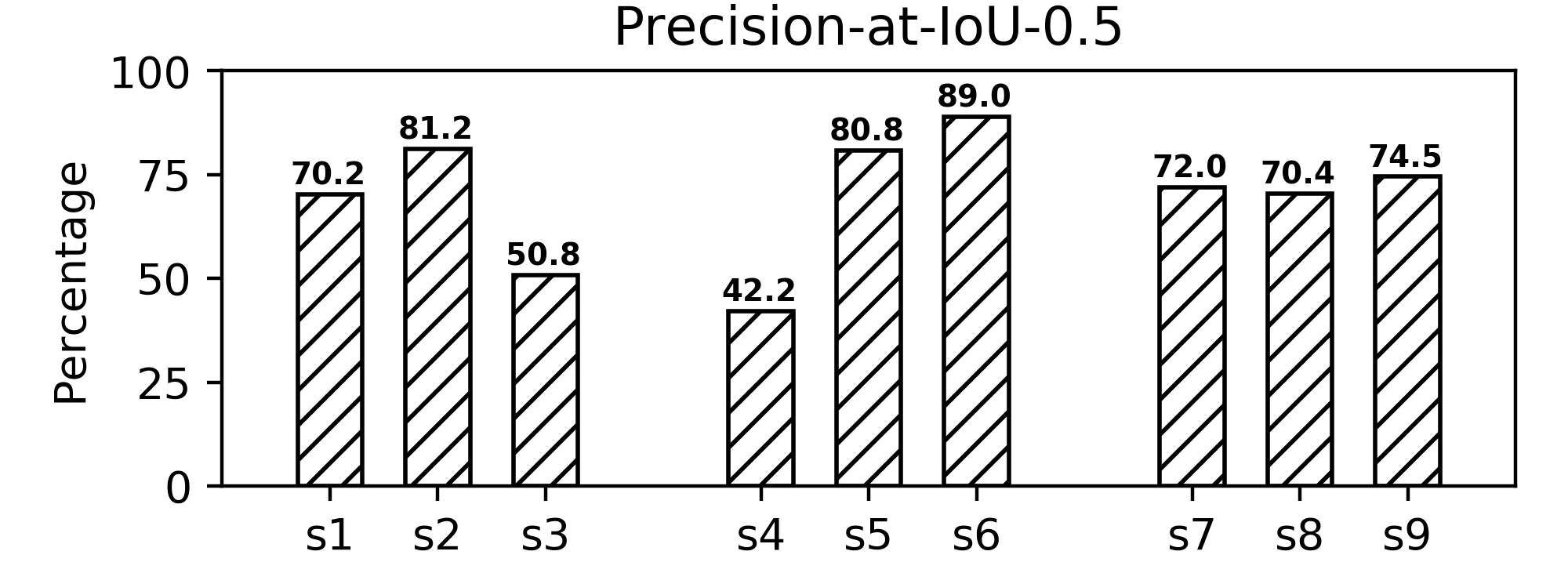}
    %\vspace{-0.3cm}
    \caption{Performance in different scenarios.}
    \label{fig:stage23ps}
    %\vspace{-0.2in}
    %\vspace{-0.3cm}
\end{figure}

\textbf{(3) Impact on the camera tilt angle.} We put the 9 scenarios into three buckets according to their camera tile angles (from low to high). We show the per-scenario results in Figure~\ref{fig:stage23ps}. We find the camera tilt angle is not a dominate factor in the performance of our algorithm. The algorithm can work well in both low tilt angle scenarios and high tilt angle scenarios.

\textbf{Failure Cases. }
\label{appendix:failure-cases}
From Figure~\ref{fig:stage23ps}, we can see the performance in scenario 3 and  scenario 4 is much lower than others. We find there are three common reasons that may cause failures. (1) The noise filter of GPS receivers, (2) Objects with similar motion, and (3) Shadows.

\begin{figure}[h]
    \centering
    %\vspace{-0.3cm}
    \includegraphics[width=\linewidth]{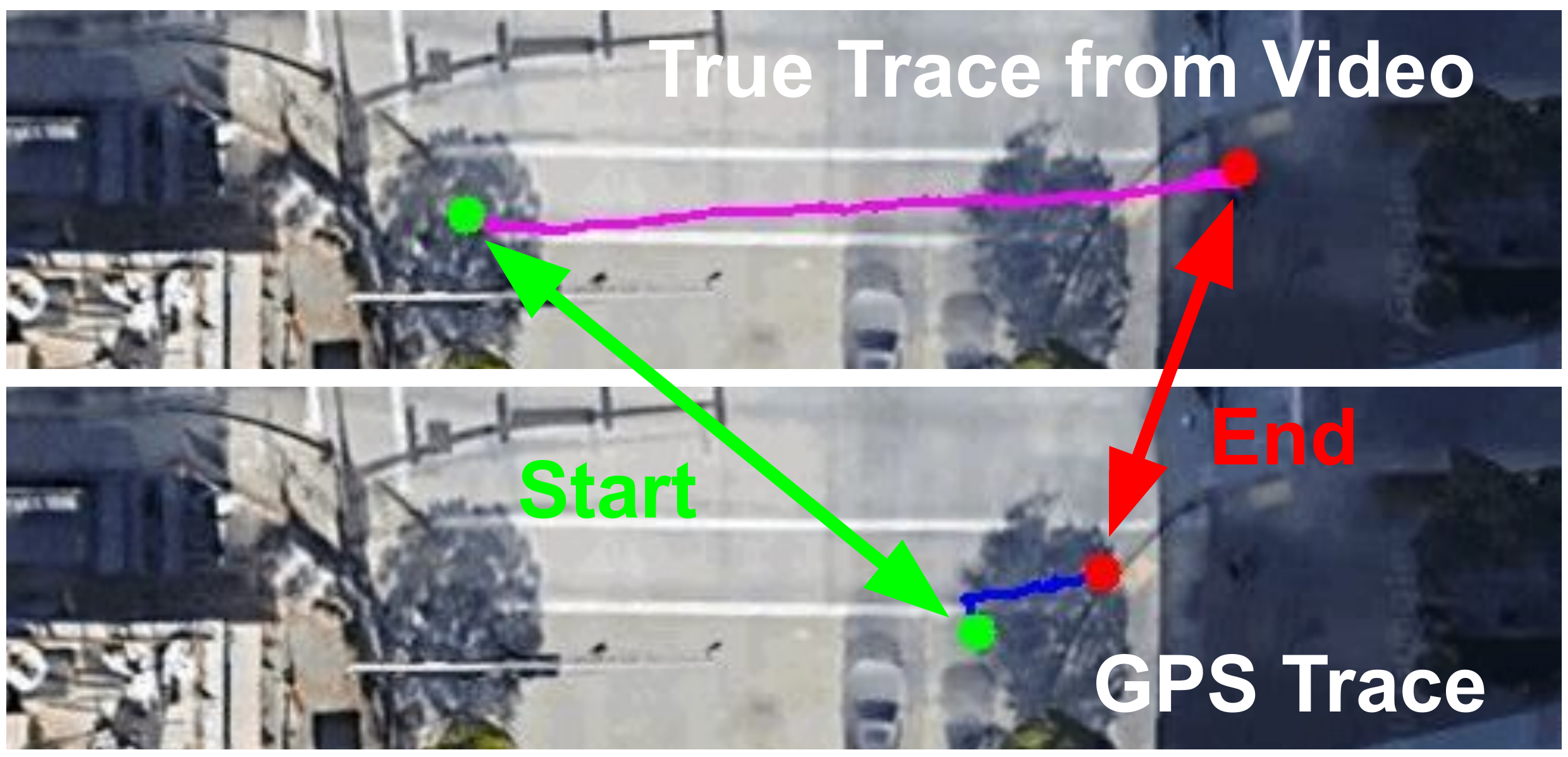}
    \caption{GPS traces and the true positions of a target object at an intersection cross-walk.}
    \label{fig:failure1}
    %\vspace{-0.3cm}
\end{figure}

\textbf{(1) GPS Noise Filter.} When the GPS signal is noisy, the de-noising algorithm may filter out the real motion of the target object. For example, in Figure~\ref{fig:failure1}, the target object moves across the road through a cross-walk. However, the GPS position stuck on a fixed location at the very beginning for a long time and failed to capture any motion of the target object. We find this behavior may lead to failures in our matching algorithm, especially when the trajectory of the target object is very short, e.g., less than 20 meters.  

%Fortunately, many GPS receivers can report their accuracy estimation, we can use this information to filter out bad GPS traces.  

\begin{figure}[h]
    \centering
    %\vspace{-0.3cm}
    \includegraphics[width=\linewidth]{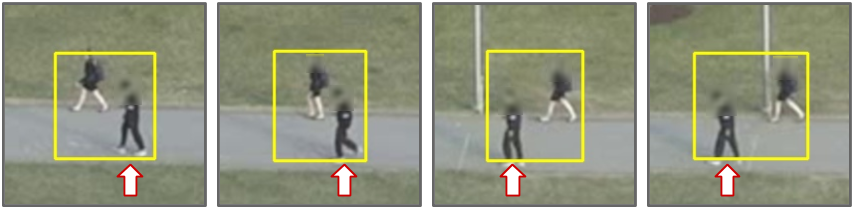}
    \caption{When there is another moving object with a similar motion to the target object (labeled by the red arrow), failure may happen. Here, the yellow rectangles are the generated bounding box annotations.}
    \label{fig:failure2}
    %\vspace{-0.3cm}
\end{figure}

\textbf{(2) Objects with similar motion.} 
We find another failure case is caused by other moving objects with a similar motion to the target object. We show an example in Figure~\ref{fig:failure2}. 

\begin{figure}[h]
    \centering
    %\vspace{-0.3cm}
    \includegraphics[width=\linewidth]{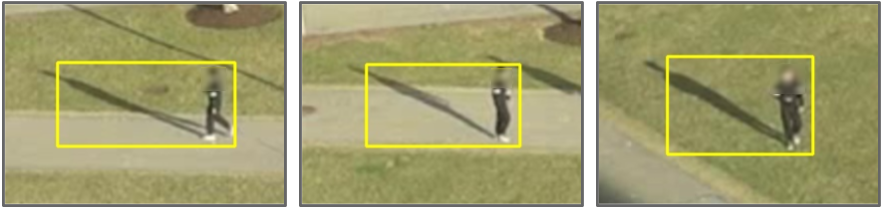}
    \caption{The shadow of the target object could be considered as a part of the target object because they have the same motion.}
    \label{fig:failure3}
    \vspace{-0.1in}
    %\vspace{-0.3cm}
\end{figure}

\textbf{(3) Shadow.} In fact, the most distracted `other' object is the shadow of the target object. We show an example in Figure~\ref{fig:failure3}, where the target object and its shadow are considered as one object. 

The reason behind this failure type and the previous one is because of the clustering algorithm used in the bounding box proposal stage. The clustering algorithm relies on the distance metric among optical flows to create objects. When two objects are close to each other for a long time, they will be treated as one object, e.g., the examples in Figure~\ref{fig:failure2} and Figure~\ref{fig:failure3}. 

We find many of the above errors can be partially detected through bounding box ranking (Section~\ref{sec:bbrank}). Meanwhile, TagMe is a proof-of-concept research prototype, we believe there are still optimization spaces left in each stage of the TagMe pipeline.

\subsection{Bounding Box Ranking}
\label{sec:bbr}

\begin{figure}[h]
    \centering
    %\vspace{-0.3cm}
    \includegraphics[width=0.9\linewidth]{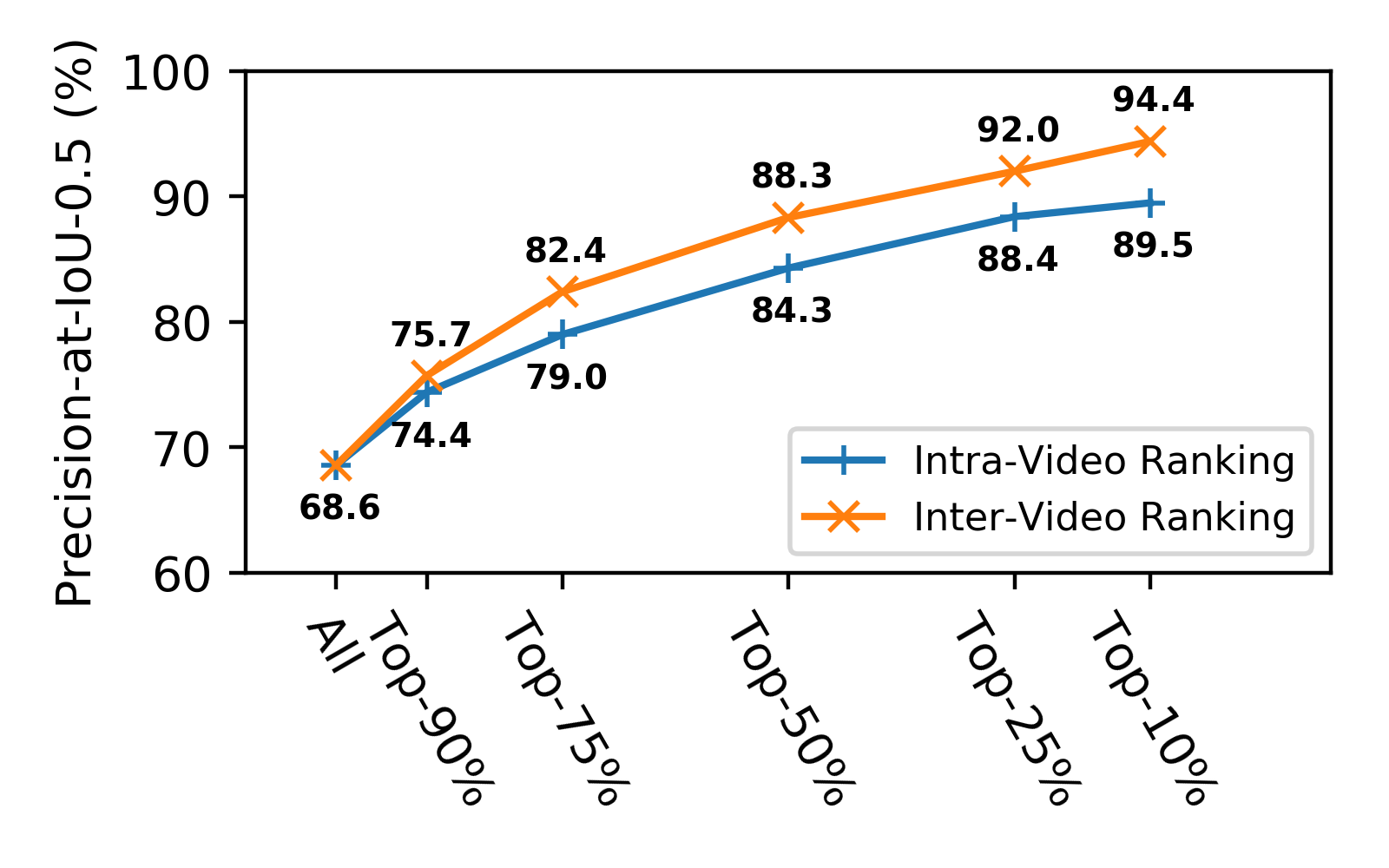}
    
    %\vspace{-0.3cm}
    \caption{Bounding box ranking purifies the annotations.}
    \label{fig:stage4}
    \vspace{-0.2cm}
\end{figure}

We evaluate the bounding box ranking performance with two ranking policies, intra-video ranking and inter-video ranking. With intra-video ranking, we rank the bounding boxes in each video clip separately. With inter-video ranking, we rank the bounding boxes generated from all videos together. In Figure~\ref{fig:stage4}, we show the quality of the remaining bounding boxes when we filter out the low-rank bounding boxes using the two ranking policies.  

We find that the bounding box quality prediction model works as expected. For example, if we only consider the top-50\% of the generated bounding boxes, the precision increases from 68.6\% to 84.3\% with the intra-video ranking policy, and to 88.3\% with the inter-video ranking policy. Comparing the intra-video ranking policy with the inter-video ranking policy, we find the inter-video ranking policy can purify the bounding boxes more efficiently. This is because inter-video ranking policy can avoid excluding too many high-quality bounding boxes in a video where most of the bounding boxes have good quality, and avoid including too many bad quality bounding boxes in a video where most of the bounding boxes have bad quality.   

\begin{figure}[h]
    \centering
    %\vspace{-0.3cm}
    \includegraphics[width=\linewidth]{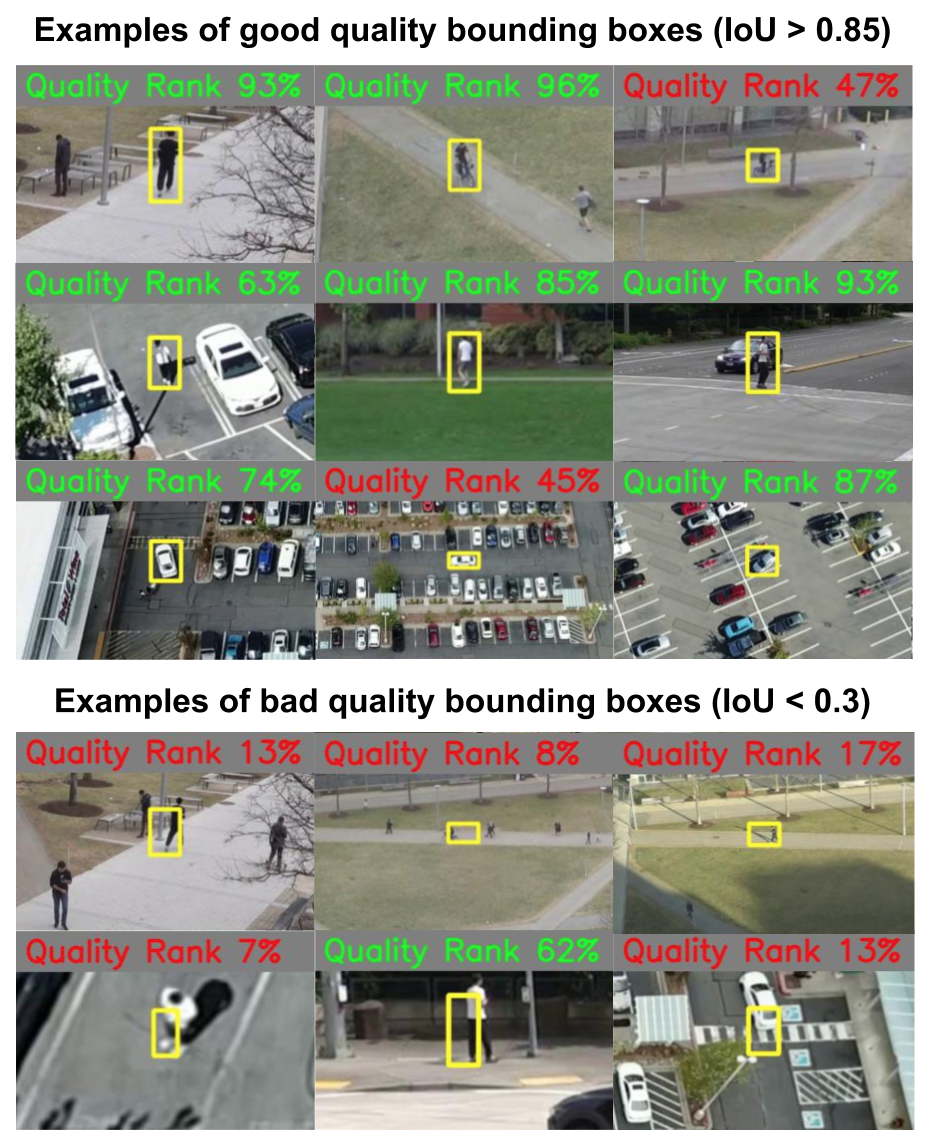}
    %\vspace{-0.3cm}
    \caption{Examples of good and bad bounding boxes. We show the corresponding quality ranks in green if the quality rank is greater than 50\%, otherwise, we show them in red.}
    \label{fig:quality}
    \vspace{-0.2cm}
\end{figure}

\textbf{Qualitative Results.} Besides the quantitative evaluation, we show examples of good bounding boxes (IoU > 0.85) and bad bounding boxes (IoU < 0.3), as well as their inter-video quality ranks in Figure~\ref{fig:quality}. These examples are random samples from our dataset. We can find the quality prediction is a good indicator for bounding box quality - most of the good quality bounding boxes have high quality ranks while most of the bad quality bounding boxes have low quality ranks.

% \subsection{Qualitative Results} 
% Please see Appendix~\ref{appendix:qualitative-results} for details.

% We show the result in Figure~\ref{fig:stage4} (the intra-video ranking curve). We find the bounding box quality prediction does its job. For example, if we only consider top-50\% of the generated bounding boxes, the precision-at-IoU-0.5 increases from 69.5\% to 84.3\%. 

% Intra-video ranking has its limitation. It may remove too many good quality bounding boxes in a video where most of its bounding boxes have good quality. To avoid this, we can use inter-video ranking, where we rank the bounding boxes generated from all videos together. 

% To enable the inter-video ranking, the predicted quality scores have to be comparable among different videos. We find the quality prediction generated by our neural network model has this property (see Appendix~\ref{appendix:inter-video-ranking}). We can use this property to perform inter-video ranking. As shown in Figure~\ref{fig:stage4}, inter-video ranking can purify the bounding boxes more efficiently. When we only consider the top-50\% of the generated bounding boxes, using inter-video ranking can improve the precision-at-IoU-0.5 from 84.3\% to 88.3\%.

\subsection{\name's Annotation Expense}

Although \name\ doesn't require human operators, it still consumes computing resources. Here, we measure the running time of processing all the 100 video clips on an AWS c5.9xlarge instance~\cite{aws}. We show the total running time and the running time of each pipeline component in Table~\ref{table:runtime}. We find that the motion analysis and the object proposal components are the two most resource-intensive components that make up 98\% of the total CPU time. The HMM matching and the refinement \& ranking components are much faster. 

\begin{table}[h]
	\centering
	\begin{tabular}{|l|c|c|} 
		\hline
		Components &Implementation&CPU Time\\ \hline
		Motion Analysis &Python+OpenCV & 15.49 min \\
		Object Proposal &Golang & 7.90 min \\
		HMM Matching &Golang & 0.26 min\\
	    Refine \& Rank &Python+Keras & 0.16 min \\ \hline 
		\multicolumn{2}{|r|}{Total CPU time (divided by 36 cores) } & 23.81 min  \\ \hline 
		\multicolumn{2}{|r|}{Wall-clock time} & 29.44 min  \\ \hline \hline
		\multicolumn{2}{|r|}{Cost} & \textbf{76 cents}  \\ \hline 
	\end{tabular}
	\caption{The cost and the running time of processing 100 video clips on an AWS c5.9xlarge instance (36 cores, \$1.54 per hour). }
	\label{table:runtime}
	\vspace{-0.3cm}
\end{table}
We use the wall-clock time to estimate the cost of the computation. The on-demand price of the AWS c5.9xlarge instance is \$1.54 per hour~\cite{aws}, so the total cost is 76 cents. 
%Here, because we process the 100 video clips in parallel, there exist laggard processes that make the wall time (29.44 min) longer than the total CPU time (23.81 min). If we process more videos, the laggard-effect gets relieved, therefore, the per-unit-cost will become cheaper. 

We compare \name's annotation cost with the annotation cost from Google Data Labeling Service~\cite{googledatalabel}. In this service, the labeling process for our dataset consists of two tasks: (1) the \textit{video object detection task}, which only needs to label one object per video as we only have one target object in each video, and (2) the \textit{video tracking task}, which tracks the target object and generate the bounding boxes over the entire video.

In Table~\ref{table:cost}, we show the estimated labeling cost of recruiting one annotator to label the data from Google Data Labeling Service. When we only use the top-x\% of the \name's auto-annotations, we change the manual annotation workload accordingly.
%When we only consider the top-50\% or the top-10\% (from bounding box ranking) of the \name's auto-annotations, we change the estimated cost in Google data labeling service accordingly. 
In this comparison, we find that \name\ can reduce the annotation cost by one or two orders of magnitute, by a factor ranging from 18x to 110x. 

\begin{table}[h]
    \centering
	\begin{tabular}{|l|c|c|c|} 
		\hline
		 &Manual Annotation&TagMe& Saving\\ \hline
	All& \$84.24 & \$0.76 & 110x \\
	Top-50\%& \$45.12 & \$0.76 & 59x \\
	Top-10\%& \$13.82 & \$0.76 & 18x \\ \hline
	\end{tabular}
	\caption{Annotation cost reduction from \name.}
	%\name\ reduces the annotation cost by one to two orders of magnitude. 
	\label{table:cost}
	%\vspace{-0.3cm}
\end{table}

These cost savings are a conservative estimate as we usually need to recruit two or three annotators to ensure labeling quality, and the computing cost of \name\ can be much cheaper if we port the Python implementation to a faster language and use reserved instances rather than on-demand instances on AWS.

In this evaluation, we did not directly compare the annotation quality through the IoU metric or ND metric. Instead, we evaluate the annotation quality by comparing the performance of the models trained with different annotations. We show an evaluation of this in Section~\ref{sec:casestudy}. We find the quality of \name's auto-annotations (top-50\%) is similar to the manual annotations.  

%\subsection{Extension to indoor scenarios}

\section{Case Studies}
\label{sec:casestudy}
%In this Section, we show how \name\ improves a YOLO-v3 object detector (pre-tained on COCO dataset) in two scenarios.

\subsection{Per-Camera Fine-Tuning for Cyclist Detection}

We use the YOLO-v3 object detector to find cyclists from videos recorded by a surveillance camera (scenario 5 in the dataset). 
The YOLO-v3 object detector was pre-trained on the COCO dataset (80 different objects). We find that using this pre-trained model to detect cyclists in scenario 5 is difficult because of the \textit{limited object categories in the training dataset}. Although the COCO dataset covers 80 different objects, it doesn't include cyclists. We can only use the detection of bicycles to represent cyclists. As a result, the pre-trained model doesn't perform well in this scenario. To make the object detector work, we can fine-tune the model with additional annotated dataset from the target camera. In this case, we can use \name\ to collect the dataset. We find the dataset collected by \name\ is sufficient for this task. 

\begin{figure}[h]
    \centering
    %\vspace{-0.3cm}
    \includegraphics[width=\linewidth]{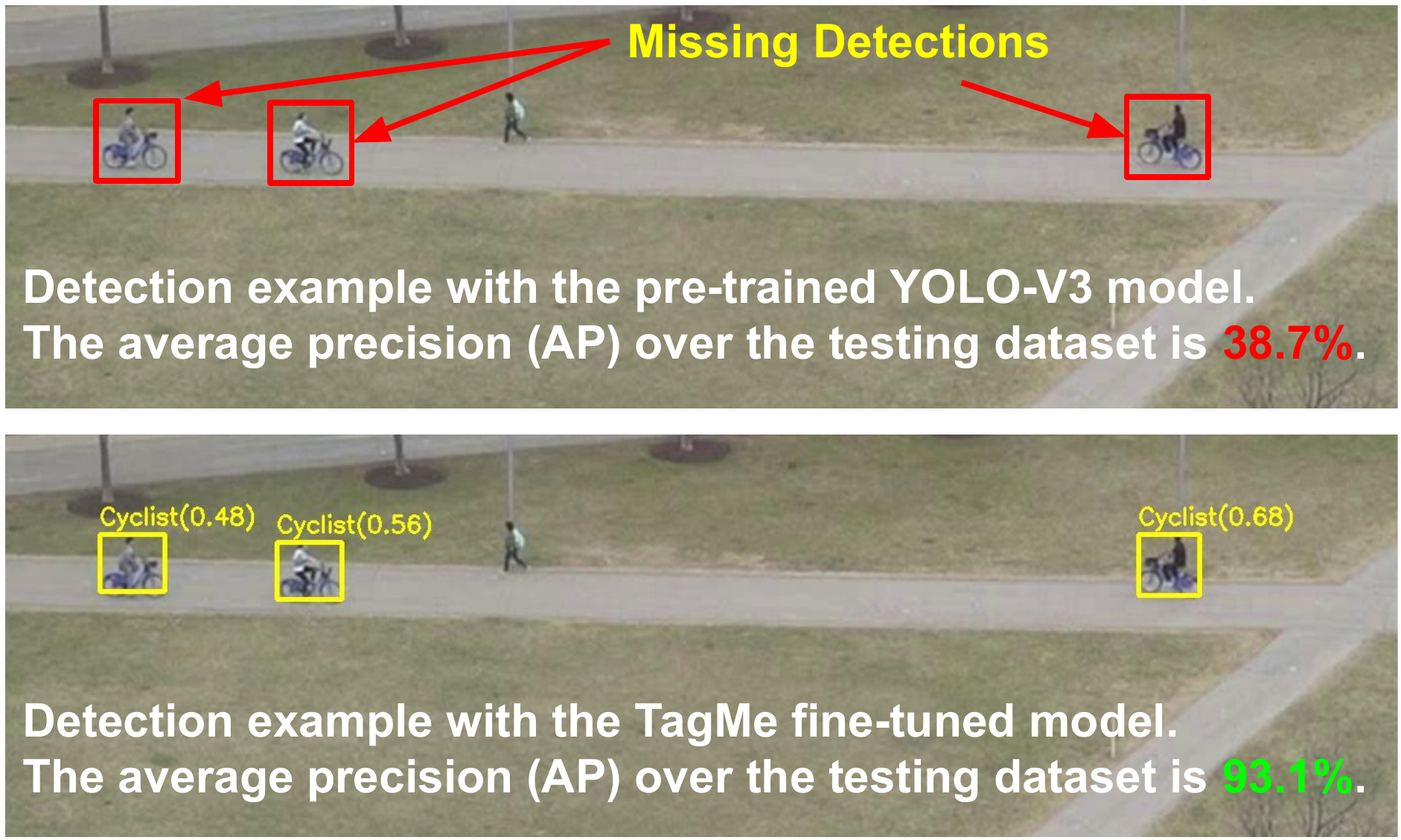}
    %\vspace{-0.1cm}
    %\includegraphics[width=\linewidth]{Figures/stage4cor2.png}
    \caption{Case Study 1: \name's auto-annotations improve the precision (AP) of cyclist detection from 38.7\% to 93.1\%.}
    \label{fig:casestudy1}
    \vspace{-0.1in}
\end{figure}

%We conducted an quantitative evaluation to prove this.  
In the evaluation, we randomly split the videos in scenario 5 into two even-sized groups. We use the videos from one group as the testing dataset % and manually label a subset of the frames, i.e., one frame per 5 seconds. 
and feed the videos from another group to \name . We use the top-50\% of the generated bounding boxes as the training dataset to fine-tune the pre-trained model. We report the average precision (AP) metric~\cite{everingham2010pascal} on the testing dataset with both the pre-trained model and the fine-tuned model (ours). We find \name's automatically-generated annotations have sufficient quality and they can significantly improve the precision of the cyclist detector, i.e., improve the AP score from 38.7\% to 93.1\%. We show detection examples from both the pre-trained model and the fine-tuned model in Figure~\ref{fig:casestudy1}. 

\subsection{Improving Person Detection in Aerial Imagery}
\label{sec:casestudy2}
\begin{figure}[h]
    \centering
    %\vspace{-0.3cm}
    \includegraphics[width=1.0\linewidth]{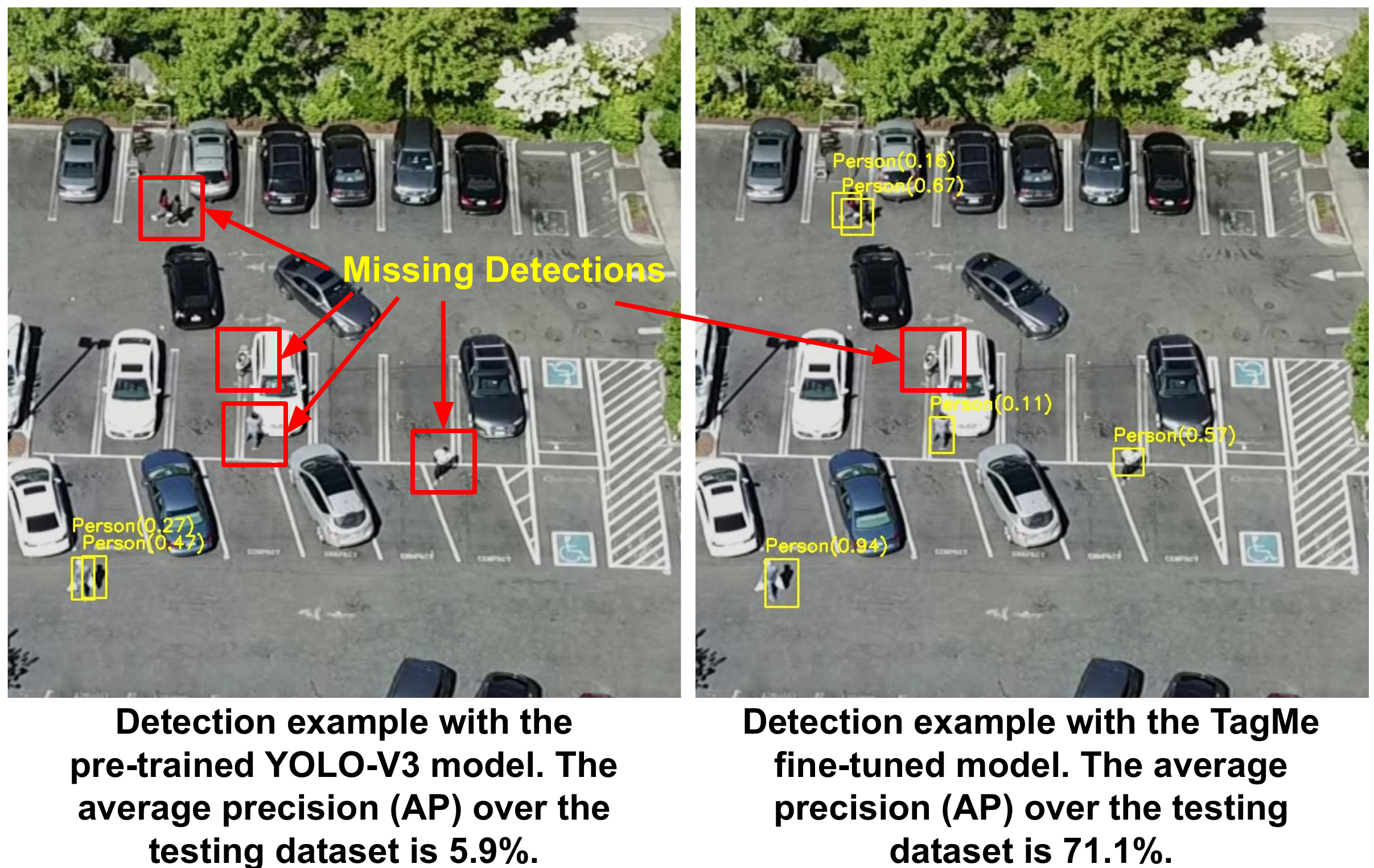}
    %\vspace{-0.1cm}
    %\includegraphics[width=\linewidth]{Figures/stage4cor2.png}
    \caption{Case Study 2: \name's auto-labels improve the precision (AP) of person detection in aerial imagery from 5.9\% to 71.1\%.}
    \label{fig:casestudy2}
    %\vspace{-0.1in}
    \vspace{-0.1in}
\end{figure}

Detecting persons from aerial imagery with the pre-trained YOLO-V3 model (on COCO dataset) is difficult. Because the camera angle (top-down) is totally different from the majority of the camera angles in the training dataset. As a result, we find the pre-trained model failed to detect many persons in scenario 7 of our dataset (Figure~\ref{fig:casestudy2}).

In this case study, we use the auto-annotations from \name\ to improve person detection in aerial imagery.
\name\ can only generate bounding boxes on the target object, but there may exist many other same-category objects (other persons) in the video. 
%When this happens, the fine-tuning may yield defective results. In fact, this is a common challenge faced by the computer vision community and many solutions have been proposed~\cite{missingobjwu2018soft,missingobjxu2019missing,missingobjyang2020object,missingobjzhang2020solving}.
% cite: https://www.groundai.com/project/object-detection-as-a-positive-unlabeled-problem/1
% cite: https://arxiv.org/pdf/1806.06986.pdf
% https://openaccess.thecvf.com/content_CVPRW_2019/papers/Weakly%20Supervised%20Learning%20for%20Real-World%20Computer%20Vision%20Applications/Xu_Missing_Labels_in_Object_Detection_CVPRW_2019_paper.pdf
To overcome this issue, we use a \textit{repeated-teaching} method~\cite{missingobjxu2019missing} to perform fine-tuning.
%We first fine-tune the YOLO-V3 model with a few epoches on the ground-truth labels generated by \name.  Then we use the model to generate the person predictions on the training dataset. After this, we merge the predictions whose confidence are greater than $0.5$ with the ground-truth labels generated by \name. Later on, we repeat the whole training process on the training dataset with the new labels.
We find this method works well in our scenario and fine-tuning on \name's auto-annotations can improve the precision (AP) of person detection from 5.9\% to 71.1\% (Figure~\ref{fig:casestudy2}).

\subsection{Quality Assessment}
We compare the quality of models fine-tuned with \name's auto-annotations and the models fine-tuned with human-labeled ground truth (GT). We find the label quality of \name\ is sufficient in these two case studies (see Table~\ref{table:quality}).
\begin{table}[h]
	%\vspace{-0.2cm}
	\centering
	\begin{tabular}{|l|c|c|} 
		\hline
		& Case Study 1 & Case Study 2 \\\hline
	No fine-tuning &38.7\%& 5.9\%  \\
	Fine-tuning on GT & 95.6\% &  71.4\%  \\
	Fine-tuning on \name &93.1\%& 71.1\% \\\hline 
	\end{tabular}
	\caption{Average precision (AP) of the object detectors.}
	\label{table:quality}
	\vspace{-0.1in}
\end{table}

\section{Related Work}
\label{sec:related}
\textbf{Video Object Detection. } Existing video object detection solutions fall into two categories\footnote{There are also weak-supervised and semi-supervised approaches which can be considered as supervised and unsupervised approaches respectively.}, supervised approaches and unsupervised approaches. Supervised approaches~\cite{redmon2018yolov3,yao2019video,beery2020context,zhu2018towards,kang2016object} often have good precision and lead the state-of-the-art performance in video object detection tasks. However, they rely on high-quality annotated datasets for training. In contrast, unsupervised approaches~\cite{yao2019video} don't rely on annotated datasets and require less computing resources, but they usually have compromised precision. In this work, we adapts an unsupervised approach at the candidate object proposal stage (the first stage) so that it doesn't rely on annotated datasets. As the goal of the candidate object proposal stage is to achieve high coverage (or recall) rather than high precision, using an unsupervised approach here is sufficient\footnote{Video object detection consists of object localization and object recognition. We only need to localize objects and we don't need to recognize each object.}. 

\textbf{Video Object Annotation. }
Given the growing need for ground-truth labels, many video object annotation tools have been proposed. Because of the temporary locality in videos, human annotators don't need to label the objects in all the video frames, instead, they can only label a few key frames and let the tools to propagate the annotations to the unlabeled frames. For example, LabelME-Video~\cite{yuen2009labelme}, VATIC~\cite{vondrick2013efficiently}, and CVAT~\cite{sekachev2019computer} allow users to annotate key frames and creates annotations for the unlabeled frames through interpolation or tracking. 

Besides interpolation and tracking, the annotations can also be propagated to the unlabeled frames through an object detector. For example, Yao~\cite{yao2012interactive} and iVAT~\cite{bianco2015interactive} train an object detector incrementally on the labeled frames and apply it to the unlabelled frames to speed up the annotation task. 

When we use an incrementally-trained object detector to propagate the object annotations, the choices of the key frames to annotate are often critical to the overall annotation time. A recent work to solve a similar problem is BubbleNets~\cite{griffin2019bubblenets}, where they use a deep learning model to sort the video frames and let the annotators label the high-rank frames first to improve the annotation efficiency.% (i.e., annotate less frames).

Although these approaches proposed many techniques to reduce the annotation cost, they all require human annotators. In contrast, \name\ uses crowd-sourced GPS traces to achieve fully-automatic object annotation in videos. Therefore, \name\ can significantly reduce the annotation cost. 

\textbf{GPS-Aided Object Localization and Tracking.}
The idea of using both GPS and videos to localize and track objects has been explored in a few works. For example, Liao~\cite{liao2012incorporation} proposed a people tracking solution which first localizes the target person in videos using GPS readings and then tracks the target person using vision-based tracking. Feuerhake~\cite{feuerhake2015gps} proposed an approach to fuse the GPS readings with the object locations obtained from videos to improve the precision of object localization, e.g., improving the precision of person localization in soccer games. 

We find the basic idea of using GPS and videos together in these works are similar to \name. However, \name\ focuses on the object annotation task in a different context. As the result, the task itself is more challenging, and we have to make different design choices and introduce specific components such as the refinement and ranking stages in \name\ to overcome the challenges.

\section{Conclusion}
\label{sec:conclusion}
In this work, we propose \name, which combines GPS traces with motion analysis of videos to automatically generate bounding box annotations. We conducted a comprehensive evaluation of \name\ and show \name\ can produce good quality annotations at much lower cost than traditional human-centric methods. 

\name\ focuses on GPS sensors in outdoor settings but the approach can extend to other localization methods in indoor settings and support other annotation types, e.g., segmentation. As an example of using mobile sensing and computing to help with machine learning tasks, TagMe makes an exploration of \textit{sensor-assisted automatic data annotation}, which holds the potential to enable low-cost annotated dataset creation at scale.

% \input{paper/abstract.tex}
% \input{paper/introduction-v2.tex}
% \input{paper/motivation.tex}
% \input{paper/design.tex}
% \input{paper/implementation.tex}
% \input{paper/evaluation.tex}
% \input{paper/related.tex}
% \input{paper/discussion.tex}
% \input{paper/conclusion.tex}

% \section{BeeCluster Resource}
% The detailed design of BeeCluster is not covered by this draft, but if you are interested in it, we do have some \href{https://docs.google.com/presentation/d/1Bn4M3TDGMrml\_ztRu77dCjTCq2sRJ8UCbdI5wzGIpuA/edit?usp=sharing}{\color{blue}{\textbf{slides (click to open)}}} describing the BeeCluster API and how does BeeCluster work.

% There are some new slides about how to program with BeeCluster API (through examples) - \href{https://docs.google.com/presentation/d/1Jbbg56L_pwkGtDyS91oI8r3Q7LqYxCv3fP07B1wt6_U/edit?usp=sharing}{\color{blue}{\textbf{Beecluster Programming Reference}}}

\newpage
\bibliographystyle{acm}
\bibliography{main}

\begin{thebibliography}{10}

\bibitem{aws}
Aws ec2 on-demand pricing.
\newblock \url{https://aws.amazon.com/ec2/pricing/on-demand/}.
\newblock Accessed: 2020-10-10.

\bibitem{googledatalabel}
Google data labeling service.
\newblock \url{https://cloud.google.com/ai-platform/data-labeling/pricing}.
\newblock Accessed: 2020-10-10.

\bibitem{hyperopt}
Hyperopt.
\newblock \url{http://hyperopt.github.io/hyperopt/}.
\newblock Accessed: 2020-10-10.

\bibitem{beery2020context}
{\sc Beery, S., Wu, G., Rathod, V., Votel, R., and Huang, J.}
\newblock Context r-cnn: Long term temporal context for per-camera object
  detection.
\newblock In {\em Proceedings of the IEEE/CVF Conference on Computer Vision and
  Pattern Recognition\/} (2020), pp.~13075--13085.

\bibitem{bejiga2016convolutional}
{\sc Bejiga, M.~B., Zeggada, A., and Melgani, F.}
\newblock Convolutional neural networks for near real-time object detection
  from uav imagery in avalanche search and rescue operations.
\newblock In {\em 2016 IEEE International Geoscience and Remote Sensing
  Symposium (IGARSS)\/} (2016), IEEE, pp.~693--696.

\bibitem{hyperoptbergstra2013making}
{\sc Bergstra, J., Yamins, D., and Cox, D.}
\newblock Making a science of model search: Hyperparameter optimization in
  hundreds of dimensions for vision architectures.
\newblock In {\em International conference on machine learning\/} (2013),
  pp.~115--123.

\bibitem{bianco2015interactive}
{\sc Bianco, S., Ciocca, G., Napoletano, P., and Schettini, R.}
\newblock An interactive tool for manual, semi-automatic and automatic video
  annotation.
\newblock {\em Computer Vision and Image Understanding 131\/} (2015), 88--99.

\bibitem{ciaparrone2020deep}
{\sc Ciaparrone, G., S{\'a}nchez, F.~L., Tabik, S., Troiano, L., Tagliaferri,
  R., and Herrera, F.}
\newblock Deep learning in video multi-object tracking: A survey.
\newblock {\em Neurocomputing 381\/} (2020), 61--88.

\bibitem{dick2019deep}
{\sc Dick, K., Russell, L., Souley~Dosso, Y., Kwamena, F., and Green, J.~R.}
\newblock Deep learning for critical infrastructure resilience.
\newblock {\em Journal of Infrastructure Systems 25}, 2 (2019), 05019003.

\bibitem{DBSCANester1996density}
{\sc Ester, M., Kriegel, H.-P., Sander, J., Xu, X., et~al.}
\newblock A density-based algorithm for discovering clusters in large spatial
  databases with noise.
\newblock In {\em Kdd\/} (1996), vol.~96, pp.~226--231.

\bibitem{everingham2010pascal}
{\sc Everingham, M., Van~Gool, L., Williams, C.~K., Winn, J., and Zisserman,
  A.}
\newblock The pascal visual object classes (voc) challenge.
\newblock {\em International journal of computer vision 88}, 2 (2010),
  303--338.

\bibitem{feuerhake2015gps}
{\sc Feuerhake, U., Brenner, C., and Sester, M.}
\newblock Gps-aided video tracking.
\newblock {\em ISPRS International Journal of Geo-Information 4}, 3 (2015),
  1317--1335.

\bibitem{forney1973viterbi}
{\sc Forney, G.~D.}
\newblock The viterbi algorithm.
\newblock {\em Proceedings of the IEEE 61}, 3 (1973), 268--278.

\bibitem{griffin2019bubblenets}
{\sc Griffin, B.~A., and Corso, J.~J.}
\newblock Bubblenets: Learning to select the guidance frame in video object
  segmentation by deep sorting frames.
\newblock In {\em Proceedings of the IEEE Conference on Computer Vision and
  Pattern Recognition\/} (2019), pp.~8914--8923.

\bibitem{haklay2008openstreetmap}
{\sc Haklay, M., and Weber, P.}
\newblock Openstreetmap: User-generated street maps.
\newblock {\em IEEE Pervasive Computing 7}, 4 (2008), 12--18.

\bibitem{horn1981determining}
{\sc Horn, B.~K., and Schunck, B.~G.}
\newblock Determining optical flow.
\newblock In {\em Techniques and Applications of Image Understanding\/} (1981),
  vol.~281, International Society for Optics and Photonics, pp.~319--331.

\bibitem{jeske2013floating}
{\sc Jeske, T.}
\newblock Floating car data from smartphones: What google and waze know about
  you and how hackers can control traffic.
\newblock {\em Proc. of the BlackHat Europe\/} (2013), 1--12.

\bibitem{joshi2012survey}
{\sc Joshi, K.~A., and Thakore, D.~G.}
\newblock A survey on moving object detection and tracking in video
  surveillance system.
\newblock {\em International Journal of Soft Computing and Engineering 2}, 3
  (2012), 44--48.

\bibitem{kang2016object}
{\sc Kang, K., Ouyang, W., Li, H., and Wang, X.}
\newblock Object detection from video tubelets with convolutional neural
  networks.
\newblock In {\em Proceedings of the IEEE conference on computer vision and
  pattern recognition\/} (2016), pp.~817--825.

\bibitem{liao2012incorporation}
{\sc Liao, H.-C., Lu, C.-Y., and Shin, J.}
\newblock Incorporation of gps and ip camera for people tracking.
\newblock {\em GPS solutions 16}, 4 (2012), 425--437.

\bibitem{cocolin2014microsoft}
{\sc Lin, T.-Y., Maire, M., Belongie, S., Hays, J., Perona, P., Ramanan, D.,
  Doll{\'a}r, P., and Zitnick, C.~L.}
\newblock Microsoft coco: Common objects in context.
\newblock In {\em European conference on computer vision\/} (2014), Springer,
  pp.~740--755.

\bibitem{rabiner1986introductionHMM}
{\sc Rabiner, L., and Juang, B.}
\newblock An introduction to hidden markov models.
\newblock {\em ieee assp magazine 3}, 1 (1986), 4--16.

\bibitem{redmon2018yolov3}
{\sc Redmon, J., and Farhadi, A.}
\newblock Yolov3: An incremental improvement.
\newblock {\em arXiv preprint arXiv:1804.02767\/} (2018).

\bibitem{sekachev2019computer}
{\sc Sekachev, B., et~al.}
\newblock Computer vision annotation tool: a universal approach to data
  annotation.
\newblock {\em Intel [Internet] 1\/} (2019).

\bibitem{song2019vision}
{\sc Song, H., Liang, H., Li, H., Dai, Z., and Yun, X.}
\newblock Vision-based vehicle detection and counting system using deep
  learning in highway scenes.
\newblock {\em European Transport Research Review 11}, 1 (2019), 51.

\bibitem{vondrick2013efficiently}
{\sc Vondrick, C., Patterson, D., and Ramanan, D.}
\newblock Efficiently scaling up crowdsourced video annotation.
\newblock {\em International journal of computer vision 101}, 1 (2013),
  184--204.

\bibitem{wu2017squeezedet}
{\sc Wu, B., Iandola, F., Jin, P.~H., and Keutzer, K.}
\newblock Squeezedet: Unified, small, low power fully convolutional neural
  networks for real-time object detection for autonomous driving.
\newblock In {\em Proceedings of the IEEE Conference on Computer Vision and
  Pattern Recognition Workshops\/} (2017), pp.~129--137.

\bibitem{missingobjwu2018soft}
{\sc Wu, Z., Bodla, N., Singh, B., Najibi, M., Chellappa, R., and Davis, L.~S.}
\newblock Soft sampling for robust object detection.
\newblock {\em arXiv preprint arXiv:1806.06986\/} (2018).

\bibitem{missingobjxu2019missing}
{\sc Xu, M., Bai, Y., Ghanem, B., Liu, B., Gao, Y., Guo, N., Ye, X., Wan, F.,
  You, H., Fan, D., et~al.}
\newblock Missing labels in object detection.
\newblock In {\em CVPR Workshops\/} (2019).

\bibitem{missingobjyang2020object}
{\sc Yang, Y., Liang, K.~J., and Carin, L.}
\newblock Object detection as a positive-unlabeled problem.
\newblock {\em arXiv preprint arXiv:2002.04672\/} (2020).

\bibitem{yao2012interactive}
{\sc Yao, A., Gall, J., Leistner, C., and Van~Gool, L.}
\newblock Interactive object detection.
\newblock In {\em 2012 IEEE conference on computer vision and pattern
  recognition\/} (2012), IEEE, pp.~3242--3249.

\bibitem{yao2019video}
{\sc Yao, R., Lin, G., Xia, S., Zhao, J., and Zhou, Y.}
\newblock Video object segmentation and tracking: A survey.
\newblock {\em arXiv preprint arXiv:1904.09172\/} (2019).

\bibitem{yuen2009labelme}
{\sc Yuen, J., Russell, B., Liu, C., and Torralba, A.}
\newblock Labelme video: Building a video database with human annotations.
\newblock In {\em 2009 IEEE 12th International Conference on Computer Vision},
  IEEE, pp.~1451--1458.

\bibitem{missingobjzhang2020solving}
{\sc Zhang, H., Chen, F., Shen, Z., Hao, Q., Zhu, C., and Savvides, M.}
\newblock Solving missing-annotation object detection with background
  recalibration loss.
\newblock In {\em ICASSP 2020-2020 IEEE International Conference on Acoustics,
  Speech and Signal Processing (ICASSP)\/} (2020), IEEE, pp.~1888--1892.

\bibitem{zhao2019object}
{\sc Zhao, Z.-Q., Zheng, P., Xu, S.-t., and Wu, X.}
\newblock Object detection with deep learning: A review.
\newblock {\em IEEE transactions on neural networks and learning systems 30},
  11 (2019), 3212--3232.

\bibitem{zhu2018towards}
{\sc Zhu, X., Dai, J., Yuan, L., and Wei, Y.}
\newblock Towards high performance video object detection.
\newblock In {\em Proceedings of the IEEE Conference on Computer Vision and
  Pattern Recognition\/} (2018), pp.~7210--7218.

\end{thebibliography}

% \clearpage
% \appendix
% \input{paper/appendix-approach}
% \input{paper/appendix-evaluation}

\end{document}